\documentclass[onefignum,onetabnum]{siamonline171218}

\usepackage{lipsum}
\usepackage{amsfonts}
\usepackage{graphicx}
\usepackage{epstopdf}
\usepackage{algorithmic}
\ifpdf
  \DeclareGraphicsExtensions{.eps,.pdf,.png,.jpg}
\else
  \DeclareGraphicsExtensions{.eps}
\fi

\newsiamremark{remark}{Remark}
\newsiamremark{hypothesis}{Hypothesis}
\crefname{hypothesis}{Hypothesis}{Hypotheses}
\newsiamthm{claim}{Claim}

\headers{Stable tNNs}{E. Newman, L. Horesh, H. Avron, and M. Kilmer}

\title{Stable Tensor Neural Networks for Rapid Deep Learning\thanks{Submitted to the editors October 31, 2018.
\funding{This work was partially funded by IBM.}}}

\author{Elizabeth Newman\thanks{Tufts University  (\email{e.newman@tufts.edu}).}
\and Lior Horesh\thanks{IBM TJ Watson Research Center.}
\and Haim Avron\thanks{Tel Aviv University}
\and Misha Kilmer\footnotemark[2]}

\usepackage{amsopn}

\usepackage{amsmath}
\usepackage{mathtools}
\usepackage{mathabx}
\usepackage{framed}
\usepackage{subfigure}
\usepackage{float}
\usepackage{hyperref}

\usepackage{tikz}

\usepackage{arydshln} 
\usepackage[]{xcolor}

\newtheorem{example}{Example}[section]
\newtheorem{derivation}{Derivation}[section]

\newcommand{\bs}{\boldsymbol}

\newcommand{\unfold}{{\normalfont \texttt{unfold}}}
\newcommand{\fold}{{\normalfont \texttt{fold}}}
\newcommand{\bcirc}{{\normalfont \texttt{bcirc}}}

\newcommand{\Acal}{\mathcal{A}}
\newcommand{\Bcal}{\mathcal{B}}
\newcommand{\Ccal}{\mathcal{C}}

\newcommand{\Fcal}{\mathcal{F}}

\newcommand{\Ical}{\mathcal{I}}

\newcommand{\Lcal}{\mathcal{L}}

\newcommand{\Rcal}{\mathcal{R}}

\newcommand{\Wcal}{\mathcal{W}}
\newcommand{\Xcal}{\mathcal{X}}
\newcommand{\Ycal}{\mathcal{Y}}
\newcommand{\Zcal}{\mathcal{Z}}

\newcommand{\Rbb}{\mathbb{R}}

\usepackage{pgfkeys} 	
\usepackage{ifthen} 		

\pgfkeys{
/tensor/.cd, 
dim1/.initial = 1,		dim1/.get = \dimOne,		dim1/.store in = \dimOne,
dim2/.initial = 1,		dim2/.get = \dimTwo,		dim2/.store in = \dimTwo,
dim3/.initial = 1,		dim3/.get = \dimThree,	dim3/.store in = \dimThree,
xshift/.initial = 0, 	xshift/.get = \xShift, 		xshift/.store in = \xShift,
yshift/.initial = 0, 	yshift/.get = \yShift, 		yshift/.store in = \yShift,
xspec/.initial = 0, 	xspec/.get = \xSpec, 		xspec/.store in = \xSpec,
yspec/.initial = 0, 	yspec/.get = \ySpec, 		yspec/.store in = \ySpec,
scale/.initial = 1, 	scale/.get = \myScale, 	scale/.store in = \myScale,
fill/.initial = white, 	fill/.get = \myFill, 		fill/.store in = \myFill,
back edges/.initial = 0, 	back edges/.get = \myBack, 	back edges/.store in = \myBack,
slice type/.initial = none, 		slice type/.get = \sliceType, 			slice type/.store in = \sliceType, 
number of slices/.initial = 1, 	number of slices/.get = \nSlices, 		number of slices/.store in = \nSlices,
slice width/.initial = 1, 		slice width/.get = \sWidth, 				slice width/.store in = \sWidth, 
}

\newcommand{\tensor}[1][]{\@tensor[#1]}
\def\@tensor[#1] (#2,#3) #4; {{ 

\pgfkeys{/tensor/.cd,#1}

\def\depthScale{0.5} 

\pgfmathsetmacro{\numSlicesMinusOne}{\nSlices-1}
\pgfmathsetmacro{\numSlicesPlusOne}{\nSlices+1}

\pgfmathsetmacro{\sliceLength}{\myScale*\dimOne}

\ifthenelse{\equal{\sliceType}{lateral}}
	{
	
	\pgfmathsetmacro{\sliceWidth}{\myScale*\sWidth*0.9*\dimTwo/\nSlices}
	\pgfmathsetmacro{\sliceGap}{\myScale*\dimTwo/(\nSlices-1) - \nSlices*\sliceWidth/(\nSlices-1)}
	\pgfmathsetmacro{\sliceDepth}{\myScale*\dimThree}
	
	} 
	{
	\ifthenelse{\equal{\sliceType}{frontal}}
		{
		
		\pgfmathsetmacro{\sliceDepth}{\myScale*\sWidth*0.9*\dimThree/\nSlices}
		\pgfmathsetmacro{\sliceGap}{\myScale*\dimThree/(\nSlices-1) - \nSlices*\sliceDepth/(\nSlices-1)}
		\pgfmathsetmacro{\sliceWidth}{\myScale*\dimTwo}
	
		}
		{
		\pgfmathsetmacro{\sliceWidth}{\myScale*\dimTwo}
		\pgfmathsetmacro{\sliceDepth}{\myScale*\dimThree}
		}

	}

\def\xFront{#2 + \xShift}	
\def\yFront{#3 + \yShift}
\def\xBack{#2 + \xShift + \depthScale*\sliceDepth + \xSpec*\sliceDepth}
\def\yBack{#3 + \yShift + \depthScale*\sliceDepth + \ySpec*\sliceDepth}

\def\aFront{(\xFront, \yFront)}
\def\bFront{(\xFront, \yFront + \sliceLength)}
\def\cFront{(\xFront + \sliceWidth, \yFront + \sliceLength)}
\def\dFront{(\xFront + \sliceWidth, \yFront)}

\def\aBack{(\xBack, \yBack)}
\def\bBack{(\xBack, \yBack + \sliceLength)}
\def\cBack{(\xBack + \sliceWidth, \yBack + \sliceLength)}
\def\dBack{(\xBack+ \sliceWidth, \yBack)}

\ifthenelse{\NOT\equal{\myFill}{nofill}}
	{
	\def\tempTensor{
		\fill[\myFill!25] \bFront -- \bBack -- \cBack -- \cFront -- cycle; 
		\fill[\myFill!75] \dFront -- \dBack -- \cBack -- \cFront -- cycle; 
		\fill[\myFill!50] \aFront rectangle \cFront;  
	
		\draw \aFront rectangle \cFront; 
		\draw \bFront -- \bBack; 
		\draw \cFront -- \cBack;
		\draw \dFront -- \dBack;
	
		\draw \bBack -- \cBack;
		\draw \cBack -- \dBack;
		}
	}
	{ 

	\def\tempTensor{
		\draw \aFront rectangle \cFront; 
		
		\ifthenelse{\NOT\equal{\myBack}{0}}
		{
			\draw[dashed] \bBack -- \aBack -- \dBack;
		}{}
		
		\draw \dBack -- \cBack -- \bBack;

		\ifthenelse{\NOT\equal{\myBack}{0}}
		{
			\draw[dashed] \aFront -- \aBack;
		}{}
		
		\draw \bFront -- \bBack;
		\draw \cFront -- \cBack;
		\draw \dFront -- \dBack;
		}
	}

\ifthenelse{\equal{\sliceType}{lateral}}
	{
	\foreach\sliceCount in {0,...,\numSlicesMinusOne}
		{	
		\begin{scope}[shift ={(\sliceCount*\sliceWidth + \sliceCount*\sliceGap, 0)}]
			\tempTensor;
		\end{scope}
		}
	
	}
	{
	
	\ifthenelse{\equal{\sliceType}{frontal}}
	{
	
	\pgfmathsetmacro{\xStep}{\sliceDepth/2 + \sliceGap/2 + \myScale*\dimThree*\xSpec/(\nSlices-(1-\sWidth))}
	\pgfmathsetmacro{\yStep}{\sliceDepth/2 + \sliceGap/2 +  \myScale*\dimThree*\ySpec/(\nSlices-(1-\sWidth))}
	
	\foreach\sliceCount in {-\numSlicesMinusOne,...,0}
		{	
		
		\begin{scope}[shift = {(-\sliceCount*\xStep, -\sliceCount*\yStep)}]
			\tempTensor;
		\end{scope}
	
		}
	
	}
	{
	\tempTensor;
	}
	
	}

\node at (#2 + \dimTwo/2, #3 + \dimOne/2) {#4};

}} 

\usepackage{comment}

\begin{document}

\maketitle

\begin{abstract}

We propose a tensor neural network ($t$-NN) framework that offers an exciting new paradigm for designing neural networks with multidimensional (tensor) data.  Our network architecture is based on the $t$-product \cite{KilmerMartin2011}, an algebraic formulation to multiply tensors via circulant convolution.  In this $t$-product algebra, we interpret tensors as $t$-linear operators analogous to matrices as linear operators, and hence our framework inherits mimetic matrix properties. 
%
%
To exemplify the elegant, matrix-mimetic algebraic structure of our $t$-NNs, we expand on recent work \cite{HaberRuthotto2017} which interprets deep neural networks as discretizations of non-linear differential equations and introduces stable neural networks which promote superior generalization.  Motivated by this dynamic framework, we introduce a stable $t$-NN which facilitates more rapid learning because of its reduced, more powerful parameterization.  
Through our high-dimensional design, we create a more compact parameter space and extract multidimensional correlations otherwise latent in traditional algorithms.
We further generalize our $t$-NN framework to a family of tensor-tensor products \cite{KernfeldKilmer2015} which still induce a matrix-mimetic algebraic structure.
Through numerical experiments on the MNIST \cite{MNIST} and CIFAR-10 \cite{CIFAR10} datasets, we demonstrate the more powerful parameterizations and improved generalizability of stable $t$-NNs.

\end{abstract}


\section{Introduction}

One of the main bottlenecks in deploying deep neural networks in practical applications is their storage and computational costs. 
Recent successes in the field have resulted in network architectures with millions of parameters, and the expansive trend continues. 
While it is possible to train and deploy these networks on powerful modern clusters with state-of-the-art computational techniques, their storage,
memory bandwidth, and computational requirements make them prohibitive for small devices such as mobile phones. 
%
%

One reason that so many parameters are required is that fully-connected layers of the form 
$$
A_{j+1} = \sigma(W_j \cdot A_j + \vec{b}_j)
$$
use parameters highly inefficiently \cite{Denil2014}.  
To reduce these inefficiencies, we can develop more powerful and economical parameterizations for fully-connected layers.  
Such compressed parameter spaces naturally lead to reduced memory and computational costs and can be more amenable to distributed computation.
%
%
Furthermore, high quality parameterizations can extract more meaningful information when (relevant) data is limited.   By working with an efficient, powerful parameter space, one can create a more generalizable network.

In this paper, we propose a new neural network architecture which replaces matrices with tensors (multidimensional arrays). Thus, we call our architecture {\em tensor neural networks ($t$-NNs)}. Our $t$-NNs are based on the $t$-product introduced in \cite{KilmerMartin2011}.  Under this $t$-product formalism, tensors have been shown to encode  information more efficiently than traditional matrix algorithms in applications such as facial recognition \cite{HaoKilmer2013}, tomographic image reconstructions \cite{SoltaniKilmerHansen2016}, video completion \cite{Zhang2014}, and image classification \cite{Newman2017}. In our context, the aforementioned fully-connected layers are replaced with layers of the form 
$$
\Acal_{j+1} = \sigma(\Wcal_{j} * \Acal_j + \vec{\Bcal}_{j})
$$
where $\Acal_{j+1}, \Acal_{j}, \Wcal_{j}$ are tensors, and product $*$ is the $t$-product introduced in \cite{KilmerMartin2011}.

We explore the advantages of  processing data multidimensionally in order to better leverage the structure of the data.  We demonstrate that our tensor framework yields a reduced, yet more powerful, network parameterization.  Furthermore, our framework incorporates tensors in a straightforward and elegant way which can easily extend matrix-based theory to our high-dimensional architecture.
%


Due to the flexibility and matrix mimeticability of our proposed $t$-NNs, we can easily adapt seemingly non-trivial architectural designs of neural networks in matrix space to tensor space.  
To this end, we propose a stable multidimensional framework motivated by work in Haber and Ruthotto \cite{HaberRuthotto2017} and illustrate the topological advantages of stable forward propagation.  These advantages yield a more robust classification scheme which is essential for network generalizability.
Using stable $t$-NNs, we obtain a more efficient parameterization and hence learn a classifying function more rapidly than with an analogous matrix architectures, as illustrated by our numerical experiments.   


This paper is organized as follows.  
In \cref{sec:background and preliminaries}, we give background notation and definitions for the $t$-product as well as motivation for using tensors as we do in neural networks.  
In \cref{sec:tensor neural networks}, we derive the $t$-product layer evolution rules and tensor loss functions.  
In \cref{sec:stable neural networks}, we introduce a stable $t$-NN framework and illustrate the topological benefits of stable formulations.
In \cref{sec:general tnn}, we generalize our $t$-NNs to a family of tensor-tensor products.
In \cref{sec:numerical results}, we provide numerical support for using $t$-NNs over comparable traditional neural networks frameworks.
In \cref{sec:conclusion}, we discuss future work including implementations for higher-order data and new $t$-NN designs.

\section{Related work}

Deep neural networks have been highly successful at recognition tasks, such as image and speech recognition, and natural language processing \cite{Krizhevsky2012, LeCun1998, Hinton2012}.  However, there are many limitations to deep learning, including computational and storage costs and a lack of generality   \cite{Goodfellow2016,Papernot2016, Mhaskar2016}.

To address the computational bottlenecks, neural networks have begun to incorporate tensor-based methods.  Recent work \cite{Phan2010} has demonstrated that multidimensional approaches (e.g., tensor decompositions) can extract  more meaningful features when data is naturally high-dimensional.   Some new neural network models use \emph{tensor-train layers} to extract multidimensional information \cite{Novikov2015,Stoudenmire2016}.  
	This framework applies a sequence or \emph{train} of core matrices to each element of an input tensor and returns an element of an output tensor.
Algorithmically, this tensor-train layer reduces the number of learnable parameters in the network and can be implemented efficiently.     

A different multidimensional approach introduced in \cite{Chien2018} called  a \emph{tensor factorized neural network} applies various weight matrices to an input vector, then fuses together the output feature vectors with \emph{tensor weights}.  More specifically, the output feature vectors are applied along the different modes or dimensions of a tensor of weights.
As before, this framework greatly reduces the number of parameters while still obtaining good accuracy results.


As with other tensor approaches, we are able to reduce the number of learnable parameters in our \emph{tensor neural network ($t$-NN)} because of the maintained high-dimensional structure.  However, our approach offers \emph{matrix mimetic} virtues because we regard tensors as \emph{$t$-linear operators}.  For example, we are able to formulate natural layer evolution rules that extend traditional formulas to our multidimensional framework.   This gives us an advantage over other tensor-based approaches; we can incorporate matrix-based theory while maintaining the $t$-linear integrity (multidimensional correlations) present in naturally high-dimensional data.

Recent work \cite{HaberRuthotto2017, HaberRuthottoHolthamJun2017, He2015} has addressed the issue of characterizing performance of deep residual neural networks, thereby suggesting ways of achieving enhanced performance and making deep learning more broadly applicable. Because of our matrix-mimetic algebraic framework, we can extend some of these results under our multidimensional framework.

\section{Background and preliminaries}\label{sec:background and preliminaries}


By avoiding vectorization of the data, our $t$-NN framework has the potential to extract multidimensional correlations.
We introduce our design for third-order tensors (i.e., three-dimensional arrays), but the theory easily extends to higher dimensions \cite{Martin2013, Hao2014}.  

\subsection{Tensor preliminaries}

\begin{figure}[H]
\centering
\includegraphics[scale = 0.45]{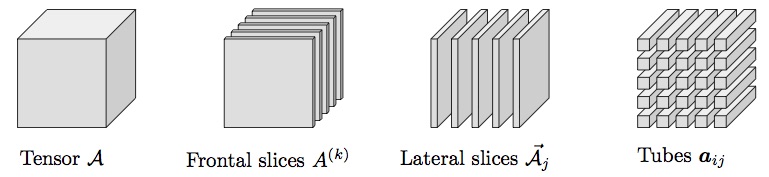}
\caption{Tensor notation.}
\label{fig:tensor notation}
\end{figure}

 Let $\Acal$ be a real-valued $\ell\times m\times n$ tensor.  Fixing the third-dimension, \emph{frontal slices} $A^{(k)}$ are $\ell\times m$ matrices for $k = 1, \dots, n$.  Fixing the second-dimension, \emph{lateral slices} $\vec{\Acal}_j$ are $\ell\times n$ matrices oriented along the third dimension for $j = 1,\dots,m$.  Fixing the first and second dimensions, \emph{tubes} $\bs a_{ij}$ are $n\times 1$ vectors oriented along the third dimension for $i = 1,\dots, \ell$ and $j = 1,\dots,n$.  We depict these divisions in \cref{fig:tensor notation}.

With this tensor notation, we introduce the \emph{$t$-product}, a method of multiplying tensors via circulant convolution, which requires the following functions:
%
%
%

\begin{definition}[Bcirc, unfold, and fold] \label{defn:bcirc}
Given  $\Acal \in \Rbb^{\ell\times m\times n}$, $\bcirc(\Acal)$ is an $\ell n\times mn$ block-circulant matrix of the frontal slices defined as follows:
	\begin{align}\label{eqn:bcirc}
	\mathtt{bcirc}(\Acal) &= \begin{pmatrix}
		A^{(1)} & A^{(n)} & \dots & A^{(2)}\\
		A^{(2)} & A^{(1)} & \dots & A^{(3)}\\
		\vdots & \vdots & \ddots & \vdots \\
		A^{(n)} & A^{(n-1)} & \dots & A^{(1)}\\
		\end{pmatrix}
	\end{align}
	
We define $\mathtt{unfold}(\Acal)$ as the first block-column of \eqref{eqn:bcirc} and $\mathtt{fold}(\mathtt{unfold}(\Acal)) = \Acal$.
\end{definition}



\begin{definition}[$t$-product \cite{KilmerMartin2011}]\label{defn:tprod}
Given $\Acal \in \Rbb^{\ell\times p\times n}$ and $\Bcal \in \Rbb^{p\times m\times n}$, the \emph{$t$-product} is defined as: 
	\begin{align}\label{eqn:tprod}
	 \Ccal = \Acal * \Bcal = \mathtt{fold}(\mathtt{bcirc}(\Acal) \cdot \mathtt{unfold}(\Bcal)), \quad \Ccal \in \Rbb^{\ell\times m\times n}.
	\end{align}
\end{definition}


For later derivations, it will be useful to consider the following $t$-product formula for a particular frontal slice:
\begin{equation}\label{eqn:tprod slice}
C^{(k)} = A^{(k)} \cdot B^{(1)} + \sum_{i=1}^{k-1} A^{(i)} \cdot B^{(k-i+1)} + \hspace{-0.15cm} \sum_{i=k+1}^n A^{(i)} \cdot B^{(n-i+k+1)}
\quad \text{ for } k = 1,\dots, n.
\end{equation}

We choose the $t$-product as our tensor operator because of its efficient implementation, though other tensor-tensor products are possible and discussed in \cref{sec:general tnn}.  It is well-known that the (normalized) discrete Fourier transform (DFT) block-diagonalizes block-circulant matrices \cite{KilmerBraman2013, HaoKilmer2013}.  This block-diagonalization amounts to taking one-dimensional Fourier transforms along the third dimension.  Thus, the $t$-product can be implemented as independent matrix multiplications in the Fourier domain as follows:
	\begin{equation}\label{eqn:tprod fourier}
	\hat{C}^{(k)} = \hat{A}^{(k)} \cdot \hat{B}^{(k)} \quad \text{ for }k = 1,\dots, n,
	\end{equation}
where $\hat{\Acal} = \texttt{fft}(\Acal,[],3)$ and $\texttt{fft}$ denotes the fast Fourier transform.  We then use the inverse Fourier transform to compute $\Ccal$; i.e., $\Ccal = \texttt{ifft}(\hat{\Ccal},[],3)$.  This algorithm is \emph{perfectly parallelizable} and therefore extremely efficient \cite{KilmerBraman2013, HaoKilmer2013}.

\subsection{Motivation: Improved parameterization with tensors}

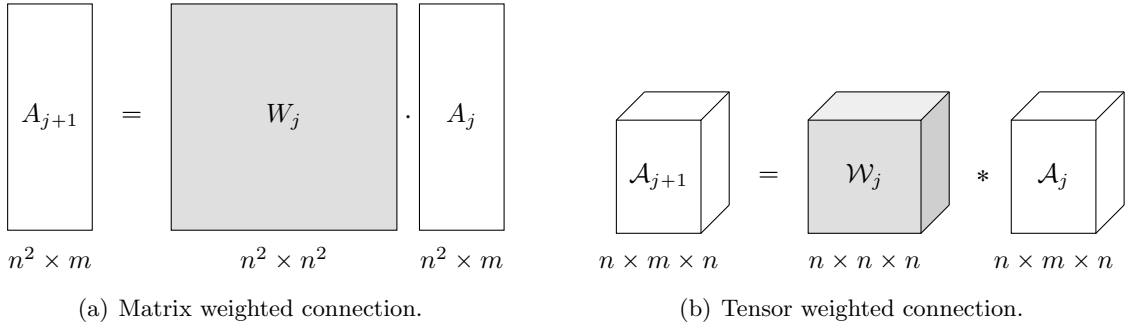
\begin{figure}[H]
\centering
\subfigure[Matrix weighted connection. \label{subfig:matrix resnet}] {
\begin{tikzpicture}[scale = 0.75]

	\draw[] (-1.4,0) rectangle (-1.4-1.5,4) node[midway] {\small $A_{j+1}$};
	\node at (-1.4-0.75,-0.5) {\small $n^2\times m$};

	\node at (-0.7,2) {$=$};

	\draw[fill = lightgray!50] (0,0) rectangle (4,4) node[midway] {\small $W_j$};
	\node at (2,-0.5) {\small $n^2\times n^2$};

	\node at (4.2,2) {$\cdot$};

	\draw[] (4+0.4,0) rectangle (4+0.4+1.5,4) node[midway] {\small $A_j$};
	\node at (4+0.4+0.75,-0.5) {\small $n^2\times m$};

\end{tikzpicture}
}
\hspace{0.5cm}
\subfigure[Tensor weighted connection. \label{subfig:tensor resnet}] {
\begin{tikzpicture}[scale = 0.75]

	\tensor[dim1 = 2, dim2 = 1.5, dim3 = 1] (-1.2-1.5,0) {\small $\Acal_{j+1}$};
	\node at (-1.2-0.75,-0.5) {\small $n\times m\times n$};

	\node at (0,1) {$=$};

	\tensor[dim1 = 2, dim2 = 2, dim3 = 1, fill = lightgray] (0.7,0) {\small $\Wcal_j$};
	\node at (0.7+1,-0.5) {\small $n\times n\times n$};

	\node at (3.5+0.3,1) {$*$};

	\tensor[dim1 = 2, dim2 = 1.5, dim3 = 1] (4+0.3,0) {\small $\Acal_{j}$};
	\node at (4+0.75+0.3,-0.5) {\small $n\times m\times n$};
\end{tikzpicture}
}

\caption{Parameterization of matrix and tensor products for fixed number of features.}
\label{fig:matrix vs. tensor resnet}
\end{figure}

One motivation for using tensors and the $t$-product in neural networks  is that we can parameterize our network to elicit more efficient feature extraction. For example, suppose we have $m$ samples of two-dimensional data of size $n\times n$.  
We can vectorize these samples and store them as columns of a matrix $A$ of size $n^2\times m$ or orient these samples as lateral slices stored in a tensor $\Acal$ of size $n\times m\times n$.
In \cref{fig:matrix vs. tensor resnet}, we compare the parameterization of the weights connecting network layers for both matrices and tensors when we fix the number of output features.

Notice that \cref{subfig:matrix resnet} requires $n^4$ weight parameters while  \cref{subfig:tensor resnet} requires only $n^3$.  Thus, the search space is smaller for tensors with the $t$-product which is computationally advantageous as the size of the data and network increases.
This efficient parameterization can be even more substantial for higher-dimensional data.   

\begin{figure}[H]
\centering
\subfigure[$W_j$ split into $n\times n$ blocks.]
	{\includegraphics[scale = 0.5]{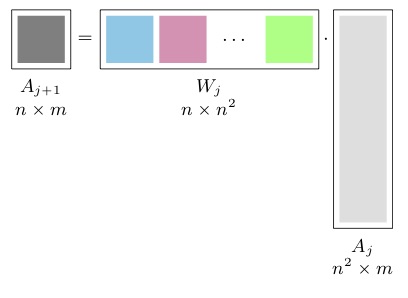}}
\hspace{0.5cm}
\subfigure[$\bcirc(\Wcal_j)$; first block-row = $W_j$.]
	{\includegraphics[scale = 0.5]{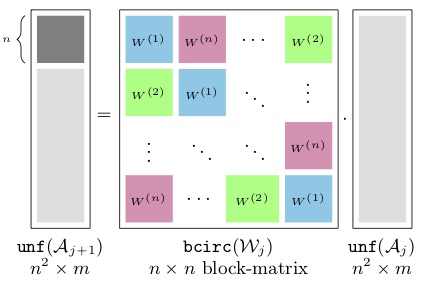}}
\caption{Featurization from matrix and tensor products for fixed number of weight parameters.}
\label{fig:improved parameterization}
\end{figure}

Alternatively, $t$-NNs can provide a more powerful parameterization for a fixed number of parameters.  Suppose we fix the number of weight parameters to be $n^3$; that is, we compare a weight matrix $W_j$ of size $n\times n^2$ to a weight tensor $\Wcal_j$ of size $n\times n\times n$.  Using \Cref{defn:tprod}, we choose the frontal slices of $\Wcal_j$ such that the first block-row of $\bcirc(\Wcal_j)$ is equivalent to $W_j$ as depicted in \Cref{fig:improved parameterization}.

In \Cref{fig:improved parameterization}, we demonstrate that for the same number of parameters, the tensor weights can capture the same features as the matrix weights and additional features from applying circulant shifts of the frontal slices.  Thus, we are able to extract more features for the same number of learnable parameters; i.e., a more powerful parameterization.

\subsection{A cohesive tensor algebra}

Among the most potent algebraic advantages of the $t$-product framework are its matrix-mimetic properties. In particular we consider tensors to be \emph{$t$-linear operators} analogous to matrices being linear operators.  Tensors act on lateral slices, thus we consider lateral slices as analogous to vectors (hence the notation $\vec{\Acal}$ in \cref{fig:tensor notation}).  For this reason, we store data as lateral slices in our framework.  To complete the analogy, tubes are the scalars of our tensor space (e.g., tubes commute under the $t$-product). For more details, see \cite{KilmerBraman2013}.

The matrix-mimetic properties of the $t$-product give rise to the following useful definitions:

	\begin{definition}[Tensor transpose]\label{defn:transpose}
Suppose $\Acal\in \Rbb^{\ell\times m\times n}$.  Then, $\Acal^{\top} \in \Rbb^{m\times \ell\times n}$ is the transpose of each frontal slice with slices $2$ through $n$ reversed.
\end{definition}

We reverse the order of the last frontal slices so that $\bcirc(\Acal^{\top}) = \bcirc(\Acal)^{\top}$.  It will be convenient to think of the order reversal in \cref{defn:transpose} as the following frontal slice mapping:
\begin{align}\label{eqn:frontal slice mapping}
(1)\mapsto (1)\quad  \text{ and }\quad (k)\mapsto (n-k+2) \quad \text{ for }k = 2,\dots,n.
\end{align}

	\begin{definition}[Identity and inverse]
	The \emph{identity tensor} $\Ical \in \Rbb^{m\times m\times n}$ is a tensor whose first frontal slice is the $m\times m$ identity matrix and the remaining slices are zero.
	
	Given $\Acal, \Bcal \in \Rbb^{m\times m\times n}$.  If $\Bcal * \Acal = \Acal * \Bcal = \Ical$, then $\Bcal$ is the \emph{inverse} of $\Acal$, denoted $\Acal^{-1}$.
	\end{definition}
	
Notice that $\bcirc(\Ical)$ is an $mn\times mn$ identity matrix, as desired.  Alternatively, we can interpret an identity tensor in terms of identity tubes.  An \emph{identity tube} $\bs e_1$ is the first standard basis vector oriented along the third dimension.  Hence, an identity tensor $\Ical$ can be defined as follows:
	\begin{align}\label{eqn:identity tensor with tubes}
	\Ical_{ii} = \bs e_1 \text{ for }i = 1,\dots, m.
	\end{align}

The tubal interpretation in \cref{eqn:identity tensor with tubes} will be useful in \cref{exam:tubal softmax}.


\section{Forming and training tensor neural networks} \label{sec:tensor neural networks}
 Suppose we have the tensors $\Acal_j \in \Rbb^{\ell_j\times m\times n}$, $\Wcal_j \in \Rbb^{\ell_{j+1}\times \ell_j\times n}$, and $\vec{\Bcal}\in \Rbb^{\ell_{j+1}\times 1\times n}$.  Then, we define tensor forward propagation as:
	\begin{equation}\label{eqn:tensor forward prop}
	\Acal_{j+1} = \sigma(\Wcal_{j} * \Acal_j + \vec{\Bcal}_{j}) \quad \text{ for }j = 0,\dots,N-1,
	\end{equation}
where $\sigma$ is an element-wise, nonlinear activation function and $N$ is the number of layers in the network.  Note that $\Wcal_{j}$ maps from layer $j$ to layer $j+1$ and $\Acal_{j+1} \in \Rbb^{\ell_{j+1}\times m\times n}$.  The summation operator $``+"$ adds  $\vec{\Bcal}_j$ to each lateral slice of $\Wcal_j * \Acal_j$.

\newcommand{\tube}{
	\begin{tikzpicture} 
	\def\n{0.1}
	\def\d{0.5}
	
	\fill[lightgray!25] (0,\n) -- (0.5*\d, \n+0.5*\d) -- (\n+0.5*\d, \n+0.5*\d) -- (\n,\n) -- cycle;
	\fill[lightgray!75] (\n,0) -- (\n+0.5*\d, 0.5*\d) -- (\n+0.5*\d, \n+0.5*\d) -- (\n,\n) -- cycle;
	
	\draw (0,\n) -- (0.5*\d, \n+0.5*\d) -- (\n+0.5*\d, \n+0.5*\d) -- (\n,\n);
	\draw (\n,0) -- (\n+0.5*\d, 0.5*\d) -- (\n+0.5*\d, \n+0.5*\d);
	
	\draw[fill = lightgray!50] (0,0) rectangle (\n,\n);
	
	\end{tikzpicture}}

Typically, we apply a classification matrix at the last layer of our network to reshape the output to the target matrix size.  
We generalize this approach to tensors by applying a \emph{classification tensor} $\Wcal_N\in \Rbb^{p\times \ell_N\times n}$ to the final layer of the network, where $p$ is the number of classes.  
	
\subsection{Tensor loss function}

Our goal is to minimize the following objective function $\bs \Fcal$:
\begin{align}\label{eqn:objective}
\bs \Fcal \equiv \bs \Lcal(f(\Wcal_{N} * \Acal_{N}), C), 
\end{align}
	where $C$ is the true classification and the objective function is composed of the following:
	\begin{align}
	&\bs \Lcal: \Rbb^{p\times m} \to \Rbb & &\text{tensor loss function}\\
	&f: \Rbb^{p\times m\times n} \to \Rbb^{p\times m} && \text{scalar tubal function}
	\end{align}
	 
	 Depending on the loss function $\bs \Lcal$, the true classification $C$ will be represented in different ways.  For example, if $\bs \Lcal$ is the least-squares function, then $C$ is a $p\times m$ matrix where $C_{ij} = 1$ if the $j^{th}$ sample belongs to the $i^{th}$ class.  However, if $\bs \Lcal$ is the cross-entropy function, then $C$ is a $1\times m$ vector where $C_j = i$ if the $j^{th}$ sample belongs to the $i^{th}$ class.
	
	To simplify our derivation, we define our objective function $\bs \Fcal$ in terms of only a loss function $\bs \Lcal$.  However,  we acknowledge that more complex objective functions $\bs \Fcal$ are easily realized, such as incorporated parameter constraints and regularization.  
	
	To understand our tensor loss function, we introduce two related functions on tensors: \emph{tubal functions} and \emph{scalar tubal functions}.
	

\subsubsection{Tubal and scalar tubal functions}
	
	As the name suggests, a \emph{tubal function} is a function which is applied a tensor \emph{tube-wise} (the tensor analogy to element-wise). To apply a function tube-wise, we consider the action of tubes $\bs a, \bs b \in \Rbb^{1\times 1\times n}$ under the $t$-product, which is equivalent to the action of a circulant matrix on a vector (see \cref{defn:tprod}); that is,
		\begin{align}\label{eqn:tprod tube action}
		\bs a * \bs b  \equiv  \texttt{circ}(\bs a) \cdot \texttt{vec}(\bs b),
		\end{align}
	where $\texttt{circ}(\bs a)$ is an $n \times n$ circulant matrix formed from elements of $\bs a$ and $\texttt{vec}(\bs b)$ is the $n\times 1$ vector of elements of $\bs b$.  Because of the matrix-representation in \cref{eqn:tprod tube action}, applying a tubal function is equivalent to applying a matrix function to the circulant matrix.
	
	It is well-known that matrix functions act as scalar-valued functions on the eigenvalues of a matrix.
	%
	Because the (normalized) DFT matrix diagonalizes circulant matrices, 
	applying a matrix function $h$ to a circulant matrix can be computed using the following formula:
		\begin{align}\label{eqn:tprod function apply eigenvalues}
		\underset{\text{tubal function}}{h(\bs a)} 
			\equiv \underset{\text{matrix function}}{h(\texttt{circ}(\bs a))} 
			= F\cdot h(\texttt{diag}(\hat{\bs a})) \cdot F^{-1},
		\end{align}
	where $F$ is the $n\times n$ DFT matrix and $\hat{\bs a}$ are the Fourier coefficients of $\bs a$.  Thus, applying a tubal function under the action of the $t$-product is equivalent to applying scalar-valued function element-wise in the frequency domain, and then transforming back to the spatial domain.  For more information about applying functions to tensors, see \cite{Lund2018}.

	\paragraph{\bf Scalar tubal functions}
	A scalar tubal function $f:\Rbb^{1\times 1 \times n} \to \Rbb$ is a composition of a tubal function and a scalar-valued function. More specifically, consider a tube $\bs a \in \Rbb^{1\times 1\times n}$.  Then, $f(\bs a)$ can be expressed as follows:
		\begin{align}
		f(\bs a) = g(h(\bs a)),
		\end{align}
	where $h:\Rbb^{1\times 1\times n} \to \Rbb^{1\times 1\times n}$ is a tubal function and $g:\Rbb^{1\times 1\times n}\to \Rbb$ is a scalar-valued function (e.g., the sum of the elements of the tube).  
	
	Both tubal functions and scalar tubal functions are applied tube-wise to a tensor; that is, if $\Acal \in \Rbb^{p\times m\times n}$, then $f(\Acal)\in \Rbb^{p\times m}$ where $f(\Acal)_{ij} = f(\bs a_{ij})$.

	\begin{example}[Scalar tubal softmax function]\label{exam:tubal softmax}
	\normalfont
	Given a vector $\vec{x} \in \Rbb^{p\times 1}$, the softmax function $f: \Rbb^{p\times 1} \to \Rbb^{p\times 1}$ is applied as follows:
		\begin{align}\label{eqn:vector softmax}
		f(\vec{x})_i = \frac{e^{x_i}}{\sum_{j=1}^p e^{x_j}} \quad \text{ for } i = 1,\dots, p.
		\end{align}
	The output of the softmax function is a $p\times 1$ vector whose elements are positive and sum to $1$, which can be usefully interpreted as a \emph{vector of probabilities} \cite{Nielsen2017}.  If we apply $f$ to a matrix $X\in \Rbb^{p\times m}$, we apply the function column-wise; that is, $f(X)_i = f(\vec{x_i})$ where $\vec{x}_i$ is the $i^{th}$ column of $X$.
	
	For tensors, we interpret a lateral slice $\vec{\Xcal} \times \Rbb^{p\times 1 \times n}$ ``vector of tubes,'' depicted below:
		\begin{align}\label{eqn:vector of tubes}
		\vec{\Xcal} = \begin{pmatrix} \bs x_1 \\ \bs x_2 \\ \vdots \\ \bs x_p\end{pmatrix}
				= \begin{pmatrix}  \tube \\ \tube \\ \vdots \\ \tube \end{pmatrix}
		\end{align}
		
	where $\bs x_i$ is the $i^{th}$ tube of $\vec{\Xcal}$ (i.e., $\vec{\Xcal}_i = \bs x_i$).  From this interpretation, the scalar tubal softmax function $f:\Rbb^{p\times 1\times n} \to \Rbb^{p\times 1\times n}$ is compose of the following tubal function $h$ and scalar function $g$:
		\begin{align}
		h(\vec{\Xcal})_i &= (\sum_{j=1}^p \texttt{exp}(\bs x_j))^{-1} * \texttt{exp}(\bs x_i), &&h:\Rbb^{p\times 1\times n} \to \Rbb^{p\times 1\times n}\label{eqn:softmax tubal}\\
		g(\vec{\Ycal}) &= \texttt{sum}(\vec{\Ycal}, 3), && g:\Rbb^{p\times 1\times n} \to \Rbb^{p\times 1}\label{eqn:softmax scalar}
		\end{align}
		
	where $g$ is the sum along the third dimension (i.e., sum of tubes).  Based on \cref{eqn:tprod fourier} and \cref{eqn:tprod function apply eigenvalues}, we can implement the tubal function $h$ as independent vector softmax functions \cref{eqn:vector softmax} in the frequency domain, then transform the result back to the spatial domain.
	
	 Once we have applied the tubal function $h$, we obtain a new lateral slice $\vec{\Ycal} = h(\vec{\Xcal})$.  Notice the sum of the tubes of $\vec{\Ycal}$ is the identity tube $\bs e_1$ (see \cref{eqn:identity tensor with tubes}); that is,
	 \begin{align}\label{eqn:tubal softmax sum}
	 \sum_{i=1}^p \bs y_i =  \sum_{i=1}^p  \left[(\sum_{j=1}^p \texttt{exp}(\bs x_j))^{-1} * \texttt{exp}(\bs x_i)\right]
	 	= (\sum_{j=1}^p \texttt{exp}(\bs x_j))^{-1} * (\sum_{i=1}^p \texttt{exp}(\bs x_i)) 
		=\bs e_1.
	 \end{align}

	
	Like the softmax function interpretation of a vector of probabilities, the tubal softmax function gives rise to an interpretation of  a \emph{lateral slice of tubal probabilities} (i.e., tubes that sum to the identity tube).  
	
	%
	 From \cref{eqn:tubal softmax sum}, the sum of the entries in the first frontal slice of $\vec{\Ycal}$ is equal to $1$, and the sum of the remaining frontal slices is equal to $0$.  Alternatively, if we were to sum along the the tubes of $\vec{\Ycal}$, we would return a vector of size $p\times 1$ whose entries sum to $1$.  We can interpret this vector as a vector of probabilities, analogous to the traditional softmax function.  This is the reason we define the scalar function $g$ as we did in \cref{eqn:softmax scalar}.  
	
	
	\end{example}

A softmax function or softmax layer is often used as the final layer before we evaluate our performance in a loss function $\bs \Lcal$ \cref{eqn:objective}.  Because we interpret the output of a softmax function to be a vector of probabiltiies, a cross-entropy loss function a natural choice.  Cross-entropy loss is a measure of the distance between probability distributions; in our case, the predicted distribution from the softmax output compared to the ``true" distribution from the known classification \cite{Goodfellow2016}.  The tensor cross-entropy loss function is a composition of a softmax function and a negative-log-likelihood function, as follows:
	\begin{align}\label{eqn:cross entropy}
	\bs \Lcal(X, C) =-\sum_{i=1}^m \log(x_{c_i, i}),\quad X = f(\Xcal),
	\end{align}

where $\Xcal\in \Rbb^{p\times m\times n}$ is the output tensor and $f$ is the scalar tubal softmax.  Each column of $X$, denoted $\vec{x}_i$, is a vector of probabilities for classifying the $i^{th}$ sample. The notation $c_i$ is the index of the class to which the $i^{th}$ sample belongs.  Hence, $x_{c_i,i}$ is the probability of the $i^{th}$ sample being correctly identified.  

The difference between tensor cross-entropy and traditional cross-entropy loss is the way in which we apply the softmax function $f$.  This further demonstrates the elegance of our $t$-NN framework; we can easily extend pre-existing neural network layers and loss functions to higher dimensions.

%
%
%

\subsubsection{Tensor back-propagation}
Once we evaluate out performance, we adjust the network parameters through layer evolution rules called back-propagation.  The tensor back-propagation formulas are the following:
	\begin{align}
	\delta \Acal_N &= W_N^{\top} * \partial \bs \Fcal/\partial \Acal_N\label{eqn:tensor back prop0}\\
	\delta \Acal_j &= \Wcal_{j}^{\top} * (\delta \Acal_{j+1} \odot \sigma'(\Zcal_{j+1})) \label{eqn:tensor back prop1}\\
	\delta \Wcal_j &=(\delta \Acal_{j+1} \odot \sigma'(\Zcal_{j+1})) * \Acal_{j}^{\top}\label{eqn:tensor back prop2}\\
	\delta \vec{\Bcal}_j &= \texttt{sum}(\delta \Acal_{j+1} \odot  \sigma'(\Zcal_{j+1}),2), \label{eqn:tensor back prop3}
	\end{align}
where $\delta \Acal_j := \partial \bs \Fcal/\partial \Acal_j$ is the error on the $j^{th}$ layer, $\Zcal_{j+1} = \Wcal_{j} * \Acal_j + \vec{\Bcal}_{j}$, $\sigma'$ is the derivative of the activation function,  $\odot$ is the  Hadamard element-wise product, and $\texttt{sum}(\cdot,2)$ is the sum along the second dimension (i.e., the sum of the lateral slices).  


\begin{derivation}[Tensor back-propagation]\label{proof:tensor back prop}
\normalfont
%
Our derivation relies on the formulas for matrix back-propagation described in  \cite{Rumelhart1986, Nielsen2017}.
%
The derivation of \cref{eqn:tensor back prop0} is somewhat involved and can be seen in \cref{sec:tensor loss function backprop}.  To derive \cref{eqn:tensor back prop1}, we rewrite \cref{eqn:tensor forward prop} in matrix form  (see \cref{defn:tprod}) and use the following matrix back-propagation formula:
%
%
	\begin{align}\label{eqn:backprop derivation1}
	\mathtt{unfold}(\delta \Acal_j) = \mathtt{bcirc}(\Wcal_{j})^{\top} \cdot 
		(\mathtt{unfold}(\delta \Acal_{j+1}) \odot \sigma'(\mathtt{unfold}(\Zcal_{j+1}))).
	\end{align}

From \cref{defn:transpose}, $\mathtt{bcirc}(\Wcal_{j})^{\top} = \mathtt{bcirc}(\Wcal_{j}^{\top})$, thus we can write \cref{eqn:backprop derivation1} in terms of tensors as written in \cref{eqn:tensor back prop1}.  The proof of \cref{eqn:tensor back prop3} is similar and the sum along the second dimension comes from the chain rule.

To derive \cref{eqn:tensor back prop2}, we write the forward propagation formula \cref{eqn:tensor forward prop} for a particular frontal slice as defined in \cref{eqn:tprod slice}:
	\begin{align}\label{eqn:proof frontal slices}
	A_{j+1}^{(k)} 
		&=\sigma\left(W_j^{(k)} \cdot A_j^{(1)} +  \sum_{i=1}^{k-1} W_j^{(i)} \cdot A_j^{(k-i+1)} 
			+\hspace{-0.25cm} \sum_{i=k+1}^n W_j^{(i)}\cdot A_j^{(n-i+k+1)}  + \vec{B}_j^{(i)}\right).
	\end{align}
	
Because frontal slices are matrices, we can differentiate \cref{eqn:proof frontal slices} with respect to $W_j^{(i)}$ and compute $\delta W_j^{(i)}$ as follows.  Let $\delta \tilde{\Acal}_{j+1} = \delta \Acal_{j+1} \odot \sigma'(\Zcal_{j+1})$, then:
	\begin{align}
	\delta W_j^{(i)} &= \sum_{k=1}^n 
		\frac{\partial \bs \Fcal}{\partial A_{j+1}^{(k)}} \cdot \frac{\partial A_{j+1}^{(k)}}{\partial W_j^{(i)}} \nonumber \\
		&= \delta \tilde{A}^{(i)}_{j+1} \cdot (A_{j}^{(1)})^{\top} 
	+ \sum_{k=1}^{i-1} 
			\delta \tilde{A}_{j+1}^{(k)} \cdot (A_{j}^{(n-i+k+1)})^{\top}
	 +\hspace{-0.25cm} \sum_{k=i+1}^n 
			\delta \tilde{A}_{j+1}^{(k)}  \cdot (A_{j}^{(k-i+1)})^{\top}.\label{eqn:weight backprop2}	
	\end{align}


Notice that \cref{eqn:weight backprop2} is similar to \cref{eqn:tprod slice} except for the index of the frontal slices: the first sum  contains $A^{(n-i+k+1)}$ instead of $A^{(i-k+1)}$ and the second sum contains $A^{(k-i+1)}$ instead of $A^{(n-k+i+1)}$.  This is exactly the frontal slice mapping of the tensor transpose from \cref{eqn:frontal slice mapping} and therefore we conclude:
	\begin{equation}
	\delta \Wcal_j = (\delta \Acal_{j+1} \odot \sigma(\Zcal_{j+1})) * \Acal_{j}^{\top}.
	\end{equation}


\end{derivation}


The fact that \cref{eqn:tensor back prop1}, \cref{eqn:tensor back prop2}, and \cref{eqn:tensor back prop3} are analogous to their matrix counterparts is no coincidence.  In our $t$-product framework, tensors are \emph{$t$-linear} operators \cite{KilmerBraman2013} just as matrices are linear operators.  This results in a natural high-dimensional extension of matrix-based theory, hence the mimetic simplicity of the tensor back-propagation formulas.



\section{Stability in $\bs t$-NNs}\label{sec:stable neural networks}

As the depth of a network increases (i.e., more layers), gradient-based approaches are subject to numerical instability known as the vanishing or exploding gradient problem \cite{ShalevShwartz2017, Bengio1994}.  To address this problem, Haber and Ruthotto interpret deep neural networks as \emph{discretizations of differential equations} \cite{HaberRuthotto2017, HaberRuthottoHolthamJun2017}. From this perspective, we can analyze the stability of forward propagation as well as the well-posedness of the learning problem; i.e., whether the classifying function depend continuously on the initialization of the parameters \cite{Ascher2010}.  By ensuring  stability and well-posedness,  networks will generalize better to similar data and will classify data more robustly.



\subsection{Stable tensor network architectures}


As presented for matrices in \cite{HaberRuthotto2017,HaberRuthottoHolthamJun2017}, consider the following tensor forward propagation scheme:
	\begin{align}\label{eqn:forward euler}
	\Acal_{j+1} = \Acal_j + h \cdot \sigma(\Wcal_{j} * \Acal_j +  \vec{\Bcal}_{j})\quad \text{ for } j = 0,\dots,N-1.
	\end{align}

This formula is akin to the residual network introduced in \cite{He2015} with the slight modification of a step size parameter $h$.  
We consider \cref{eqn:forward euler} to be the explicit Euler discretization of the following system of continuous differential equations \cite{HaberRuthottoHolthamJun2017}:
	\begin{align}\label{eqn:continuous ode}
	\frac{\text{d}\Acal}{\text{d}t} = \sigma(\Wcal(t) *  \Acal(t) + \vec{\Bcal}(t)) 
		\quad \text{ with } \quad  \Acal(0) = \Acal_0,
	\end{align}
over the time interval $[0,T]$.  We interpret the final time $T$ as the depth of the neural network in the discrete case.  From \cref{defn:tprod}, we can rewrite  \cref{eqn:continuous ode} in terms of matrices as follows:
	\begin{align}\label{eqn:tensor system ode}
	\unfold\left(\frac{\text{d}\Acal}{\text{d}t}\right) = \sigma(\bcirc(\Wcal(t)) \cdot \unfold(\Acal(t) + \unfold(\vec{\Bcal}))).
	\end{align}

The stability of ordinary differential equations depends on the eigenvalues of the Jacobian  $J(t)$ of the system with respect to $A$ \cite{Atkinson1989}.  Additionally, because \cref{eqn:continuous ode} is a non-autonomous system of ODEs, we require $J(t)$ to change gradually in time \cite{Ascher2010}.   

As demonstrated in \cite{HaberRuthotto2017}, this Jacobian will depend on both $\bcirc(\Wcal(t))$ and $\sigma'$.  Because  $\sigma$ is typically monotonic, the stability of \cref{eqn:continuous ode} depends on the eigenvalues of $\bcirc(\Wcal(t))$ for all $t\in [0,T]$.  We require the following related conditions to ensure a well-posed learning problem:
\begin{align}\label{eqn:stability requirement}
\underbrace{\text{Re}(\lambda_i(\bcirc(\Wcal(t)))) \le 0}_{\text{stable forward prop.}}\quad \text{ and } \quad
	\underbrace{\text{Re}(\lambda_i(\bcirc(\Wcal(t)))) \approx 0}_{\text{stable back prop.}},	
	\end{align}
for $i = 1,\dots, n^2$ and for all $t\in [0,T]$.  Because it is impractical to adjust  eigenvalues during the learning process, we introduce a forward propagation scheme which ensures well-posedness.

\subsection{Inherently stable tensor forward propagation}\label{subsec:stable forward prop}

Motivated by the presentation in \cite{HaberRuthotto2017}, we present a Hamiltonian-inspired, stable forward propagation technique for tensors.  A Hamiltonian is a system $H(\vec{a},\vec{z},t)$ which satisfies the following \cite{Brooks2011}:
	\begin{align}\label{eqn:hamiltonian}
	\frac{\text{d}\vec{a}}{\text{d}t} =  \nabla_{\vec{z}} H(\vec{a},\vec{z},t) 
		\quad \text{ and }\quad 
	\frac{\text{d}\vec{z}}{\text{d}t} = -\nabla_{\vec{a}} H(\vec{a},\vec{z},t),\quad \forall t\in [0,T].
	\end{align}
	
Inspired by physical phenomena, we can interpret $H$ as the total energy of the system separated into potential energy $T(\vec{a})$ and kinetic energy $V(\vec{z})$.   In this energy context, we interpret $\vec{a}$ as position and $\vec{z}$ as momentum or velocity.  
Hamiltonians have several nice properties including time reversibility, energy conservation, and symplecticness (see \cite{Ascher2010,Brooks2011} for details).    
%
%
For neural networks, the latter two properties preserve the topology of the data and ensure our discretization preserves well-posedness.

Consider the following symmetrized tensor Hamiltonian system \cite{HaberRuthotto2017}, written in matrix-form using \cref{defn:tprod}:
	\begin{align}\label{eqn:symmetrized hamiltonian}
	\frac{\text{d}}{\text{d}t}\begin{bmatrix} \texttt{unf}(\Acal) \\ \texttt{unf}(\Zcal)\end{bmatrix}
		=
		\sigma\left(
		\begin{bmatrix}
		0 & \hspace{-0.75cm}\bcirc(\Wcal(t)) \\ 
		-\bcirc(\Wcal(t))^{\top} &\hspace{-0.75cm} 0
		\end{bmatrix} \cdot 
		\begin{bmatrix} \texttt{unf}(\Acal(t)) \\ \texttt{unf}(\Zcal(t))\end{bmatrix} + \texttt{unf}(\vec{\Bcal}(t))
		\right),
	\end{align}

where $\Acal(0) = \Acal_0$ and $\Zcal(0) = \bs 0$.
This system  is inherently stable, i.e., independent of the choice of weight tensors $\Wcal(t)$, because of the block-antisymmetric structure of the forward propagation matrix.  Antisymmetric matrices have imaginary eigenvalues, and hence \cref{eqn:symmetrized hamiltonian} will be a system which satisfies \cref{eqn:stability requirement}.

We discretize \cref{eqn:symmetrized hamiltonian} using a leapfrog integration scheme, which is a symplectic integrator \cite{HaberRuthotto2017, Brooks2011}, and using the $t$-product as follows:
	\begin{align}\label{eqn:tensor leapfrog}
	\left\{\begin{array}{l}
	\Zcal_{j+\frac{1}{2}} = \Zcal_{j-\frac{1}{2}} - h\cdot \sigma(\Wcal_j^{\top} * \Acal_j + \vec{\Bcal}_j)\\
	\Acal_{j+1} = \Acal_j + h \cdot \sigma(\Wcal_j * \Zcal_{j+\frac{1}{2}} + \vec{\Bcal}_j)
	\end{array}\right. \quad 
	\text{ for }j = 0,\dots,N-1.
	\end{align}
	
 Because \cref{eqn:symmetrized hamiltonian} is inherently stable, the discretized analog \cref{eqn:tensor leapfrog} will be stable if the step size $h$ is small enough  and if the weights $\Wcal_j$ change gradually over the layers \cite{Brooks2011,Ascher2010}.
We illustrate the benefits of an inherently stable network in the following example.

\begin{example}[Tensor leapfrog stability]\label{exam:tensor leapfrog}
%
\normalfont
Consider a set of data in $\Rbb^3$ initialized with a mean of $0$ and a standard deviation of $3$, then labeled as follows: blue are inside a sphere of radius $3.5$, red inside a sphere of radius $5.5$, and black are outside both spheres. We train with $1200$ data points ($317$ blue, $466$ red, $417$ black) and store the data as $1\times 1\times 3$ tubes.


We forward propagate with one of the following discretizations \cref{eqn:forward prop experiment} for $N = 32$ layers.   The weights $\bs w_j$ are $1\times 1\times 3$ tubes generated randomly from a standard-normal distribution and normalized; the biases $\bs b_j$ are $1 \times 1\times 3$ tubes initialized at $\bs 0$.  
\begin{align}\label{eqn:forward prop experiment}
&\text{Forward Euler (FE)} && \text{Leapfrog} \nonumber\\
&\bs a_{j+1} = \bs a_j + h\cdot \sigma(\bs w_{j} * \bs a_{j} + \bs b_j) 
	&& \left\{\begin{array}{l}\bs z_{j+\frac{1}{2}} = \bs z_{j-\frac{1}{2}} - h \cdot \sigma(\bs w_j^{\top} * \bs y_{j} + \bs b_j)\\
 \bs y_{j+1} = \bs y_j + h \cdot \sigma(\bs w_j * \bs z_{j+\frac{1}{2}} + \bs b_j)\end{array}\right.
\end{align}
We train for $50$ epochs using batch gradient descent with a batch size of $10$ and a learning parameter of $\alpha = 0.01$.  To create smoother dynamics, we regularize the weights as follows (see \cite{HaberRuthotto2017} for details):
	\begin{align}\label{eqn:regularization}
		\bs \Rcal(\bs w) = \frac{1}{2h} \sum  ||\bs w_j - \bs w_{j-1}||_F^2 \quad \text{ for } j = 0,\dots,N-1.
	\end{align}

For simplicity, we used a least-square loss function.  We illustrate the dynamics of the various neural network discretizations in \cref{fig:tensor target}.


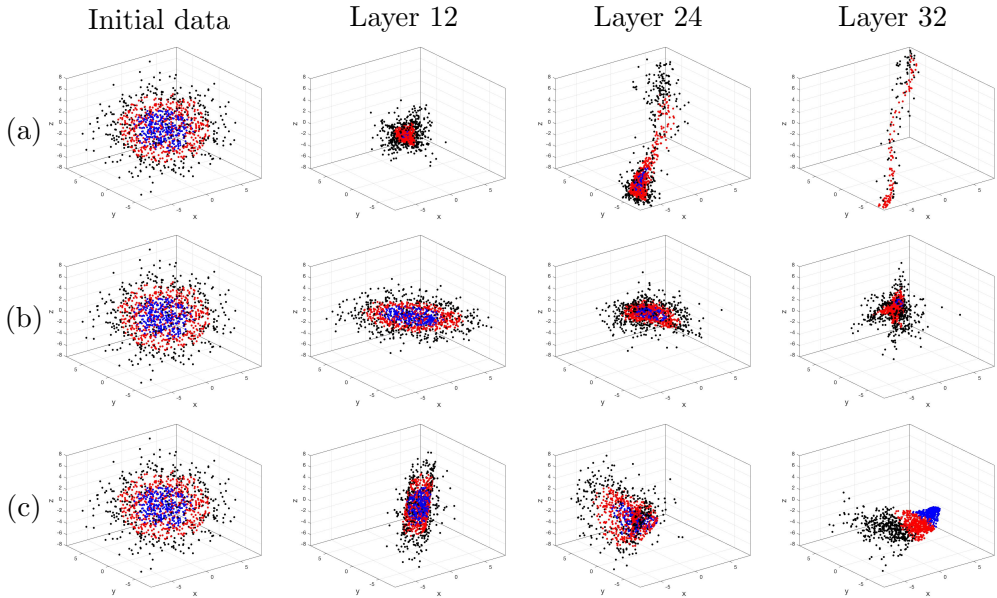
\begin{figure}[H]
\begin{tikzpicture}

\node at (0,0) (h1) {Initial data};
\node[right of = h1, node distance = 3.25cm] (h2) {Layer 12};
\node[right of = h2, node distance = 3.25cm] (h3) {Layer 24};
\node[right of = h3, node distance = 3.25cm] (h4) {Layer 32};

\node[below of = h1, node distance = 1.5cm] (fe1) 
	{\includegraphics[scale = 0.1]{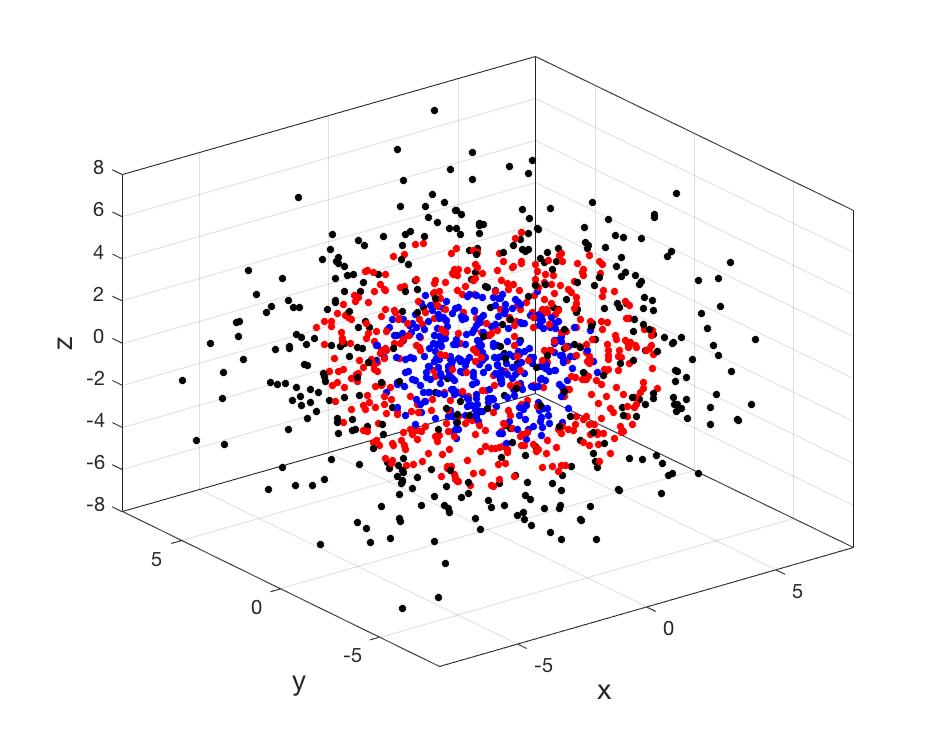}};
\node[right of = fe1, node distance = 3.25cm] (fe2) 
	{\includegraphics[scale = 0.1]{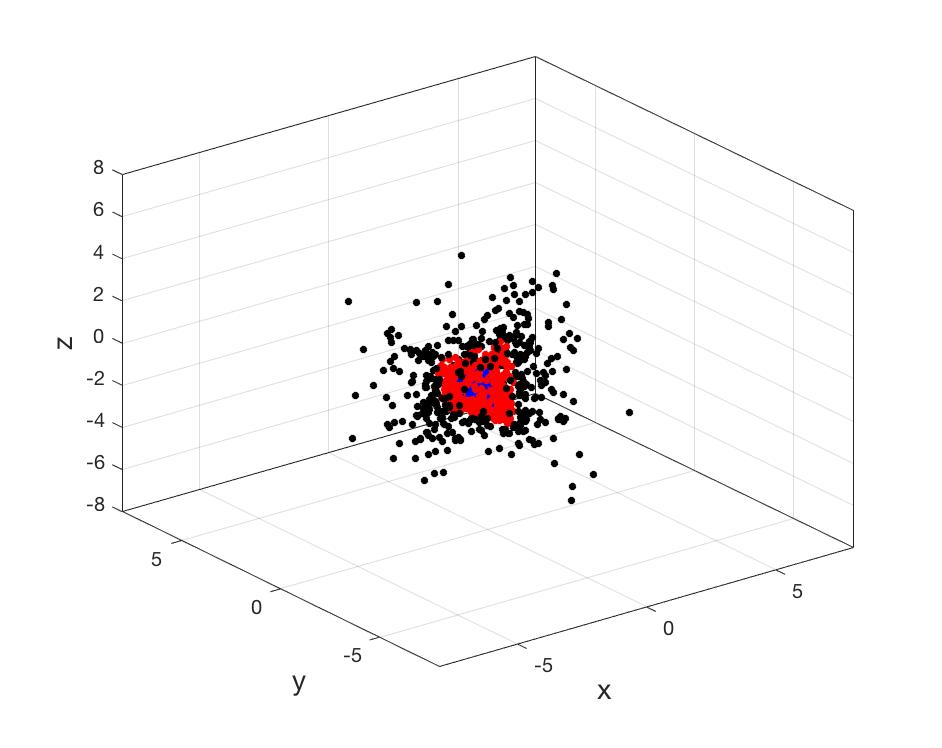}};
\node[right of = fe2, node distance = 3.25cm] (fe3) 
	{\includegraphics[scale = 0.1]{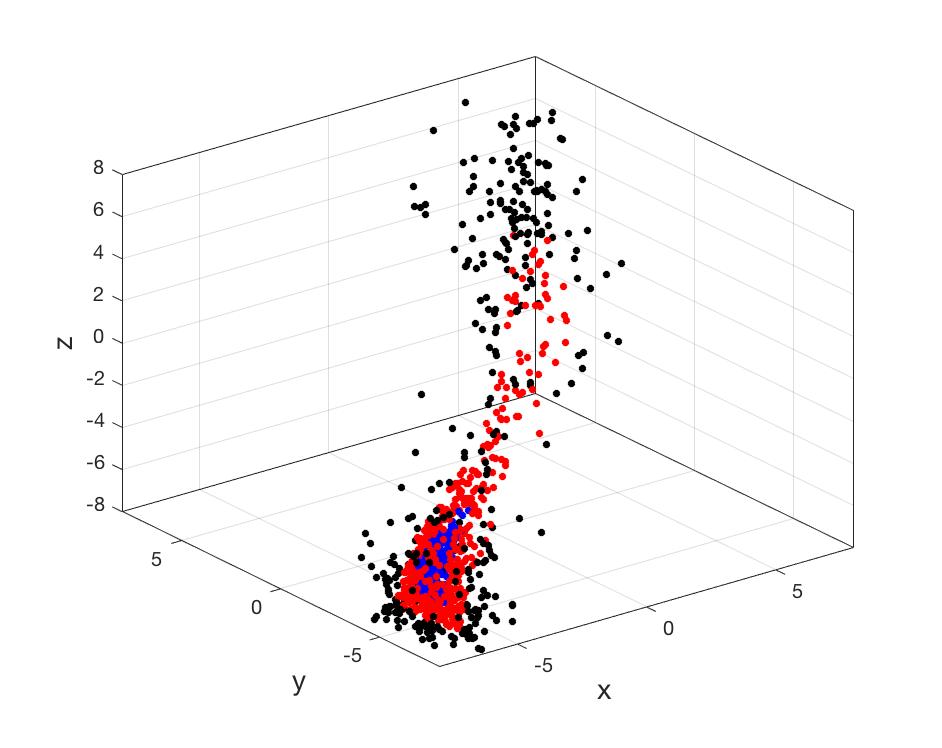}};
\node[right of = fe3, node distance = 3.25cm] (fe4) 
	{\includegraphics[scale = 0.1]{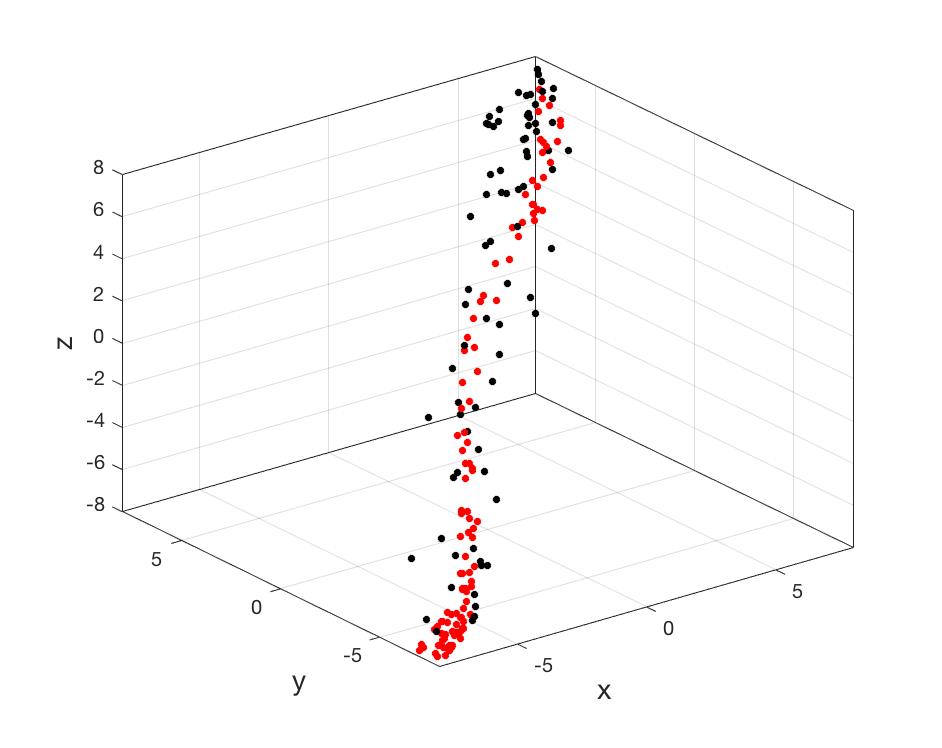}};

\node[below of = fe1, node distance = 2.5cm] (fe21) 
	{\includegraphics[scale = 0.1]{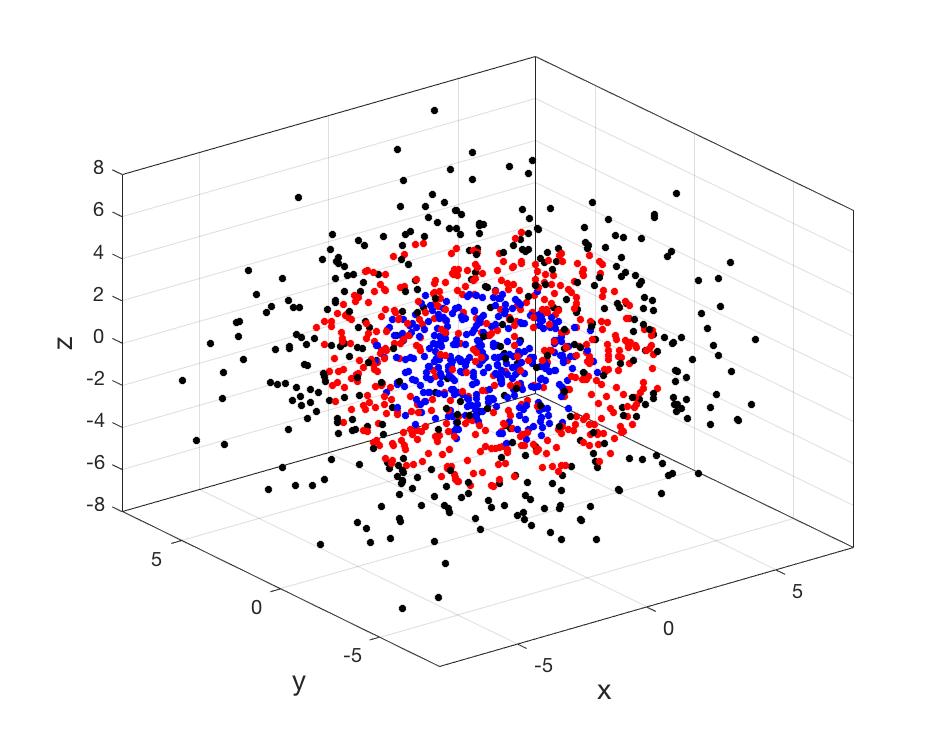}};
\node[right of = fe21, node distance = 3.25cm] (fe22) 
	{\includegraphics[scale = 0.1]{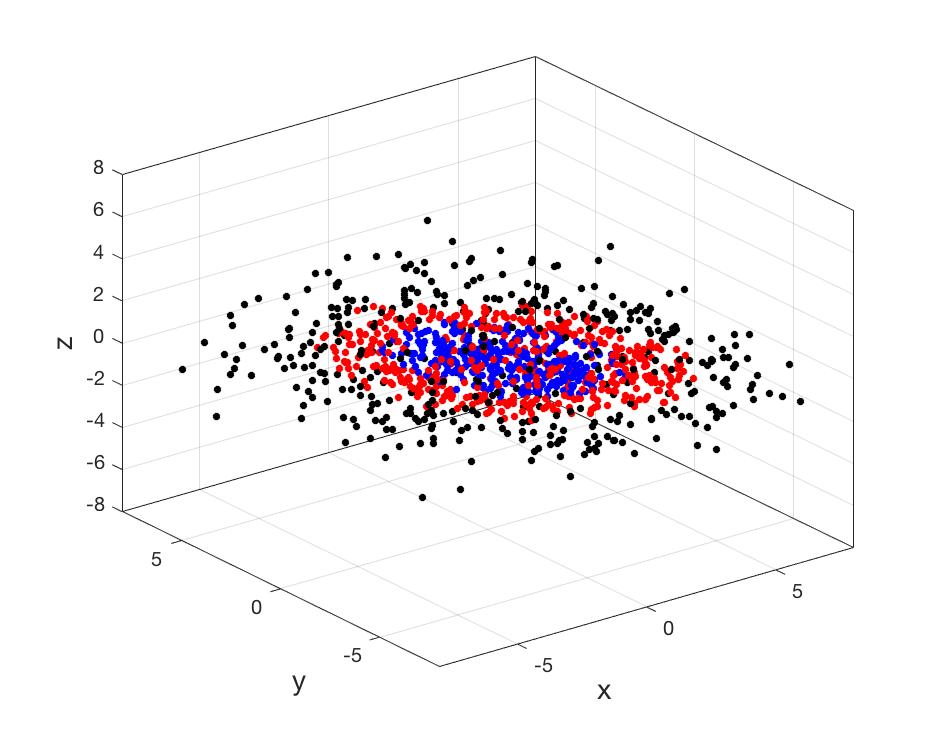}};
\node[right of = fe22, node distance = 3.25cm] (fe23) 
	{\includegraphics[scale = 0.1]{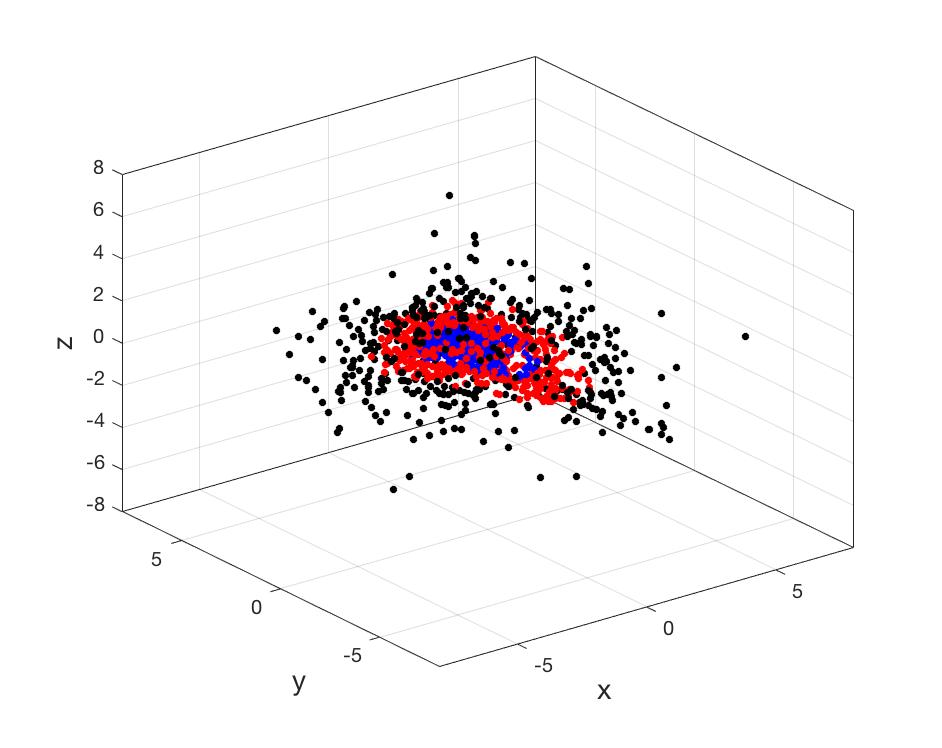}};
\node[right of = fe23, node distance = 3.25cm] (fe24) 
	{\includegraphics[scale = 0.1]{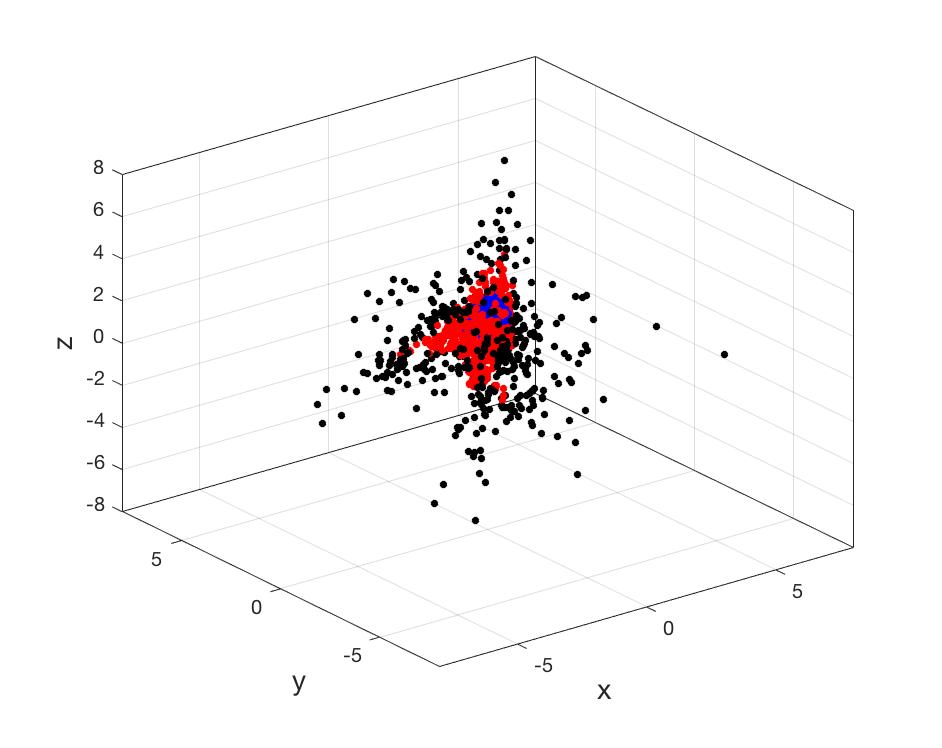}};
	
\node[below of = fe21, node distance = 2.5cm] (lf1) 
	{\includegraphics[scale = 0.1]{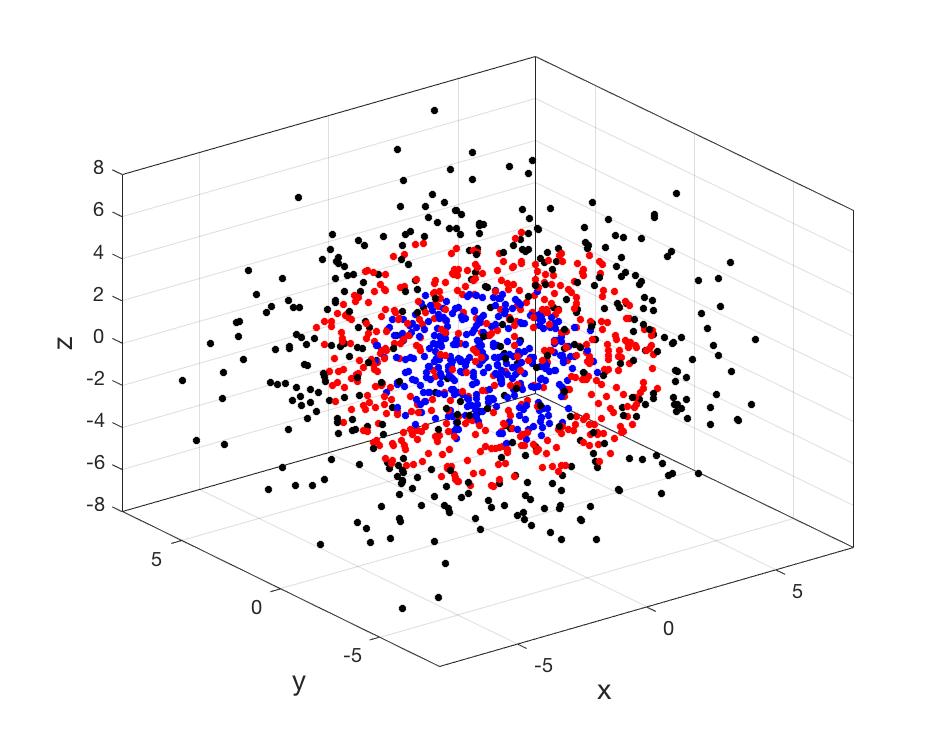}};
\node[right of = lf1, node distance = 3.25cm] (lf2) 
	{\includegraphics[scale = 0.1]{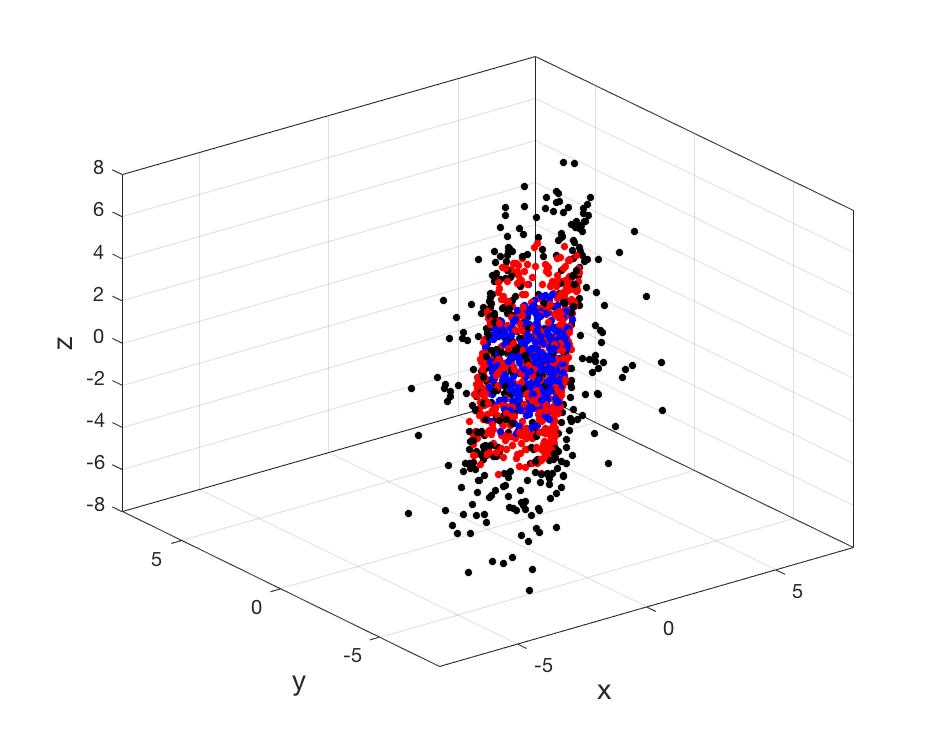}};
\node[right of = lf2, node distance = 3.25cm] (lf3) 
	{\includegraphics[scale = 0.1]{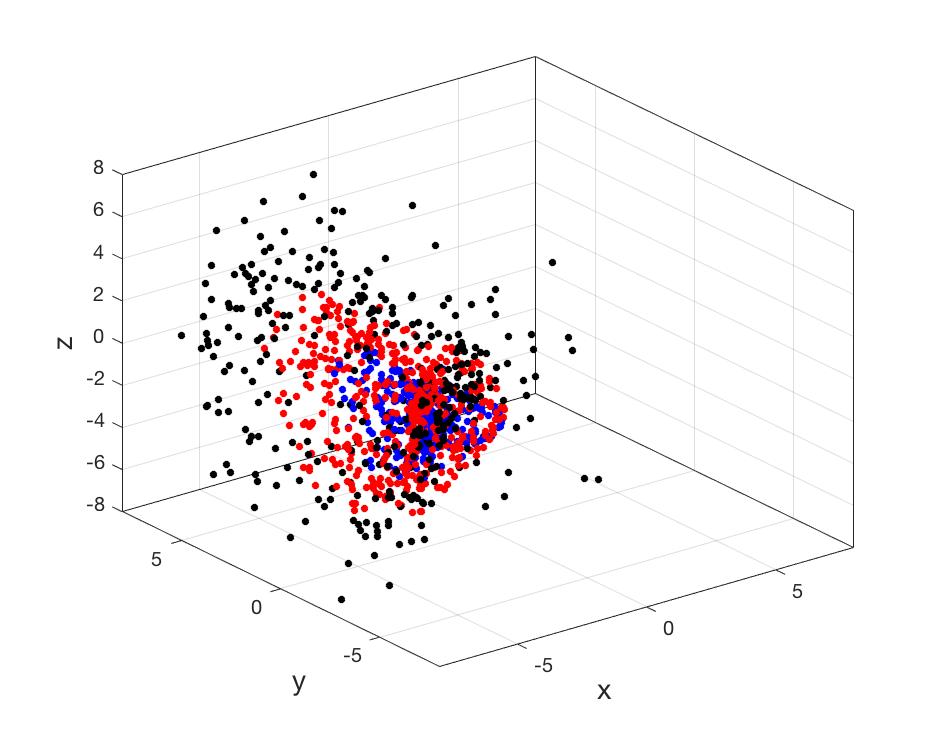}};
\node[right of = lf3, node distance = 3.25cm] (lf4) 
	{\includegraphics[scale = 0.1]{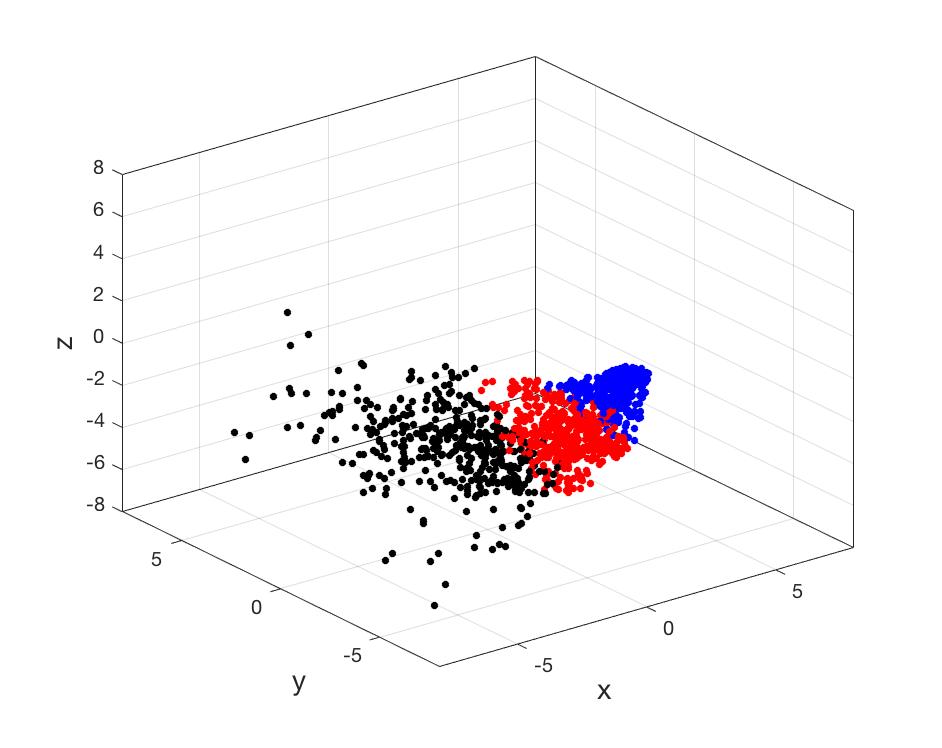}};
	
\node[left of = fe1, node distance = 1.8cm] {(a)};
\node[left of = fe21, node distance = 1.8cm] {(b)};
\node[left of = lf1, node distance = 1.8cm] {(c)};

\end{tikzpicture}

\caption{Dynamics of stable $t$-NNs: (a) FE, $h = 0.5$; (b) FE, $h = 0.25$; (c) Leapfrog, $h = 1$.}
\label{fig:tensor target}

\end{figure}


There are qualitative benefits exhibited in  \cref{fig:tensor target}(c); in particular, the output is linearly-separable by label while topologically similar to the original data.  Linear-separability is essential for accurate classification \cite{HaberRuthotto2017}, and neither forward Euler example produces a classifiable output.  Furthermore, the topology of the data changes in the forward Euler cases, such as compressing  \cref{fig:tensor target}(b) and breaking apart \cref{fig:tensor target}(a).  Such topological changes yield ill-posed learning problems and poor network generalization \cite{HaberRuthotto2017}.

\end{example}



\section{A more general $\bs t$-NN}\label{sec:general tnn}

We have demonstrated the parametric and matrix-mimetic benefits of implementing a tensor neural network based on the $t$-product; however, we can impose any tensor-tensor product within the same tNN framework.  Introduced in \cite{KernfeldKilmer2015}, the $M$-product is a tensor-tensor product based on any invertible linear transformation, and each transformation induces different algebraic properties on the space.  By forming a tNN framework under a different tensor-tensor product, we can reveal underlying correlations in the data more efficiently.  


We require a few more definitions to understand the $M$-product.
\begin{definition}[Mode-$3$ product]\label{defn:mode3}
Given $\Acal\in \Rbb^{\ell \times m\times n}$ and a matrix $M\in \Rbb^{n\times n}$, the \emph{mode-$3$ product}, denoted $\Acal \times_3 M$, is defined as
	\begin{align}\label{defn:mode3 product}
	\Acal \times_3 M = \fold_3[M \cdot \unfold_3[\Acal]],
	\end{align}
where  $\fold_3[\unfold_3[\Acal]] = \Acal$ and $\unfold_3[\Acal]\in \Rbb^{n\times \ell m}$ is a matrix whose columns are the tubes of $\Acal$.

\end{definition}

The mode-$3$ product is connected to the $\unfold$ operator (see \cref{defn:bcirc}) as follows:
	\begin{align}\label{eqn:mode3 vs unfold}
	\Acal \times_3 M = \fold((M \otimes I) \cdot \unfold(\Acal)),
	\end{align}
where $\otimes$ denotes the Kronecker product and $I$ is the $\ell\times \ell$ identity matrix.  We call the $M$ the \emph{transformation} and $\Acal\times_3 M$ is a tensor which lives in the transform space.

\begin{definition}[Facewise product] \label{defn:facewise}
Given $\Acal\in \Rbb^{\ell\times p\times n}$ and $\Bcal\in \Rbb^{p\times m\times n}$, the \emph{facewise product}, denoted $\Ccal = \Acal\smalltriangleup \Bcal$, is defined as 
	\begin{align}\label{eqn:facewise matrix}
	\Ccal = \Acal \smalltriangleup \Bcal &= \fold({\normalfont \texttt{bdiag}}(\Acal) \cdot \unfold(\Bcal))
		= \fold\left(
		{\arraycolsep=0.5pt\def\arraystretch{1}
		\begin{pmatrix} A^{(1)} &&& \\ & A^{(2)} && \\ & & \ddots & \\ &&&A^{(n)} \end{pmatrix} }
		\cdot
		\begin{pmatrix} B^{(1)} \\ B^{(2)} \\ \vdots \\ B^{(n)} \end{pmatrix}\right),
	\end{align}

where $\Ccal\in \Rbb^{\ell\times m\times n}$.  More simply, the facewise product multiplies the frontal slices of $\Acal$ and $\Bcal$ independently.
\end{definition}


With \cref{defn:mode3} and \cref{defn:facewise}, the $M$-product is defined as follows:

\begin{definition}[$M$-product]\label{defn:mprod}
Given $\Acal\in \Rbb^{\ell\times p\times n}$ and $\Bcal\in \Rbb^{p\times m\times n}$ and an invertible matrix $M\in \Rbb^{n\times n}$, the \emph{$M$-product} is defined as 
	\begin{align*}
	\Ccal = \Acal *_M \Bcal = ((\Acal \times_3 M) \smalltriangleup (\Bcal \times_3 M)) \times_3 M^{-1}, \qquad \Ccal \in \Rbb^{\ell\times m\times n}.
	\end{align*}
\end{definition}

If $M$ is the identity matrix, the $M$-product is the facewise product.  The $M$-product preserves the matrix mimetic properties that were essential for the $t$-product-based $t$-NN. 

\begin{definition}[$\bs M$-product transpose]\label{defn:mprod transpose}
Given $\Acal \in \Rbb^{\ell\times m\times n}$, the \emph{$M$-product transpose} $\Acal^\top \in \Rbb^{m\times \ell\times n}$ is the transpose of each frontal slice of the tensor.
\end{definition}

The $M$-product transpose must preserve $(\Acal \times_3 M)^\top = (\Acal^\top \times_3 M)$.

While in \cref{defn:mprod}, we define the $M$ product in terms of real-valued matrices, as demonstrated in \cite{KernfeldKilmer2015}, we can define tensor-tensor products over the complex domain.  If we allow complex-valued transformations and $M$ is the (normalized) DFT matrix, the $M$-product is equivalent to the $t$-product.  In this paper, we restrict the $M$-product to real-valued transformations to avoid potentially problematic complex matrix algebra in state-of-the-art neural network platforms like PyTorch \cite{Pytorch}.

\subsection{$\bs M$-product backpropagation}

Given an invertible transformation $M\in \Rbb^{n\times n}$, suppose we have the following forward propagation scheme:
	\begin{align}\label{eqn:mprod forward prop}
	\Acal_{j+1} = \sigma(\underbrace{\Wcal_j *_M \Acal_j + \vec{\Bcal}_j}_{\Zcal_{j+1}}) \text{ for } j=0,\dots, N-1.
	\end{align}

As before in \cref{eqn:objective}, we evaluate our performance using a tensor loss function $\bs \Lcal(f(\Wcal_N *_M \Acal_N), C)$.  If the transformation $M$ is orthogonal, the back-propagation formulas are the following:
	\begin{align}
	\delta \Acal_j &= \Wcal^\top *_M (\delta \Acal_{j+1} \odot \sigma'(\Zcal_{j+1}))\label{eqn:mprod backprop1}\\
	\delta \Wcal_j &= (\delta \Acal_{j+1} \odot \sigma'(\Zcal_{j+1})) * \Acal_j^\top \label{eqn:mprod backprop2}\\
	\delta \vec{\Bcal}_j &= \texttt{sum}(\delta \Acal_{j+1} \odot \sigma'(\Zcal_{j+1}), 2)\label{eqn:mprod backprop3}
	\end{align}

\begin{derivation}[$M$-product back-propagation]\label{deriv:mprod back prop}
\normalfont 
	It will be useful to note the following back-propagation formula for the mode-$3$ product. We differentiate a scalar-valued function $g:\Rbb^{\ell\times m\times n}\to \Rbb$ with respect to $\Acal$ as follows:
	\begin{align}\label{eqn:mode3 derivative}
	\frac{\partial}{\partial \Acal} [g(\Acal \times_3 M)] 
		&=\frac{\partial}{\partial \Acal}[g(\fold((M\otimes I) \cdot \unfold(\Acal)))] \nonumber\\
		&=\fold((M\otimes I)^\top \cdot \unfold(g'(\Acal \times_3 M))) \nonumber \\
		&= g'(\Acal\times_3 M) \times_3 M^\top.
	\end{align}
	
	For the facewise product, the derivatives are rather trivial and can be derived from the representation of in \cref{eqn:facewise matrix}.
	
%

We now have the tools we need to differentiate the $M$-product and derive our back-propagation formula as follows: 
	\begin{align}
	\frac{\partial}{\partial \Acal} [g(\Wcal *_M \Acal)] 
		&=\frac{\partial}{\partial \Acal} [g(((\Wcal \times_3 M) \smalltriangleup (\Acal \times_3 M)) \times_3 M^{-1})] \nonumber \\
		&=((\Wcal^\top \times_3 M) \smalltriangleup (g'(\Wcal *_M \Acal) \times_3 M^{-\top})) \times_3 M^\top.\label{eqn:mprod derivative}
	\end{align}

While \cref{eqn:mprod derivative} is not a compact formula, if we restrict $M$ to be orthogonal (i.e., $M^{-1} = M^\top$), we get the following elegant, matrix-mimetic formula:
	\begin{align}
	\frac{\partial}{\partial \Acal} [g(\Wcal *_M \Acal)] = \Wcal^\top *_M g'(\Wcal *_M \Acal).
	\end{align}

If the scalar-valued function $g$ is the objective function $\bs \Fcal$ in \cref{eqn:objective},  we get exactly the back-propagation formula we expected in for the error $\delta \Acal_j$ in \cref{eqn:mprod backprop1}.
Similar derivations can be used to verify the formulas for the weight and bias updates in \cref{eqn:mprod backprop2} and \cref{eqn:mprod backprop3}.

\end{derivation}


\section{Numerical results}\label{sec:numerical results}

\begin{figure}[H]
	\centering
	\subfigure[MNIST accuracy.]{\includegraphics[width=\textwidth]{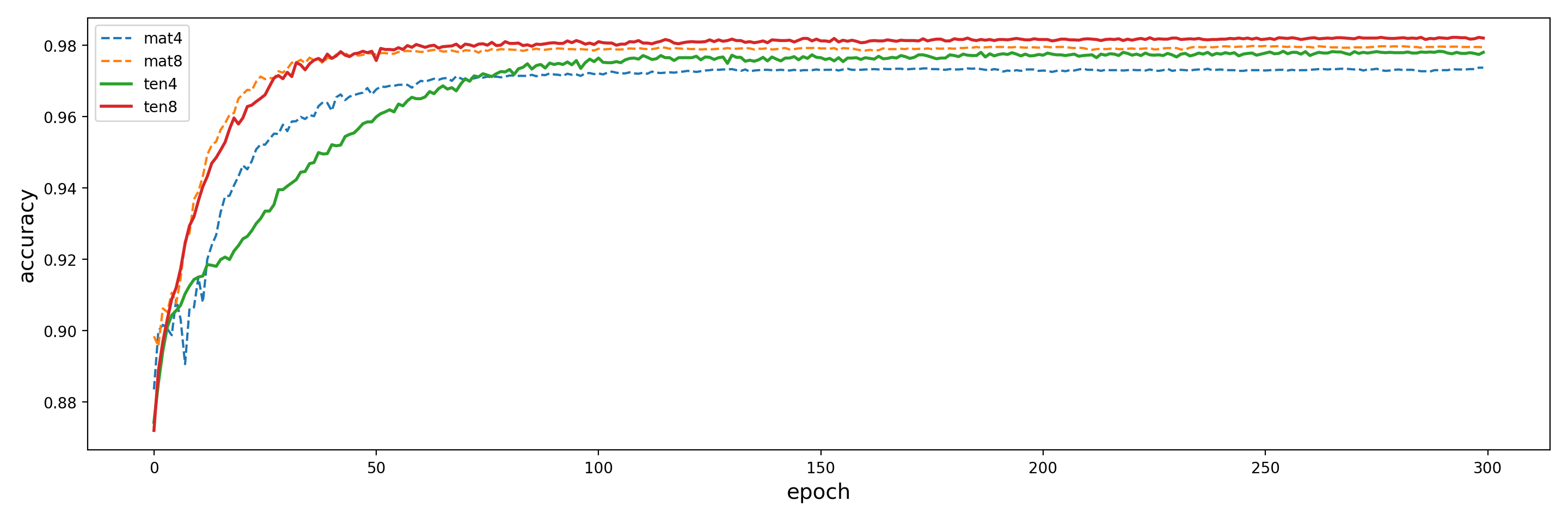}}
	
	\subfigure[MNIST loss.]{\includegraphics[width=\textwidth]{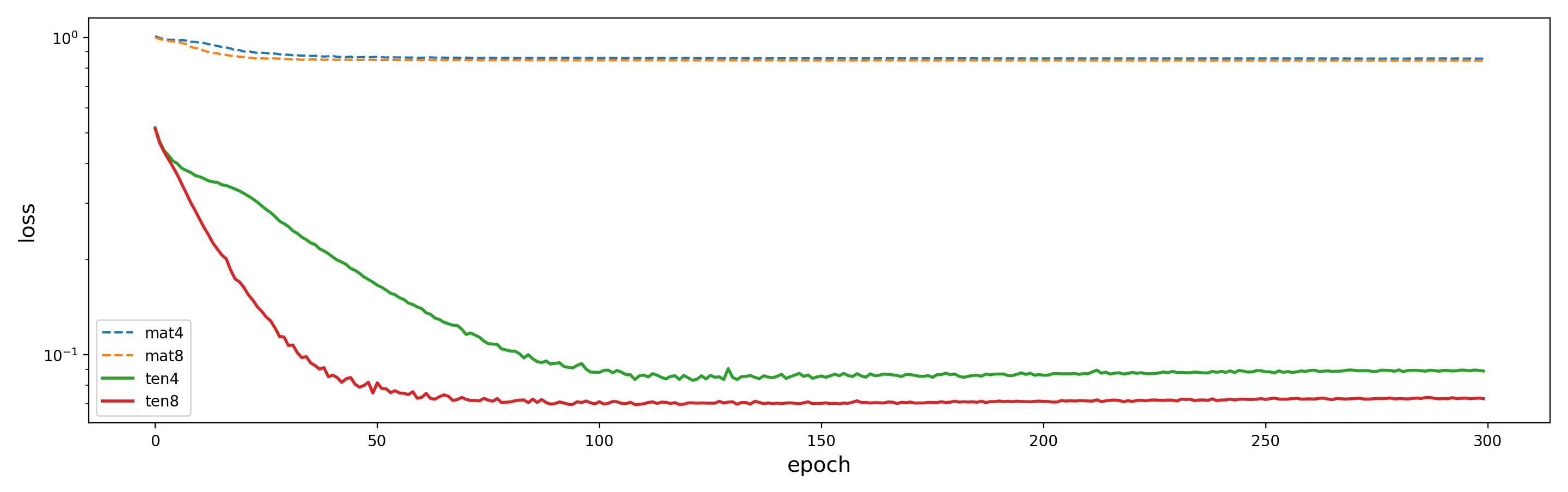}}
	\caption{MNIST results comparing matrix and tensor networks containing $4$ and $8$ leapfrog layers.}
	\label{fig:mnist results}
\end{figure}

\begin{figure}[H]
	\centering
	\subfigure[CIFAR-10 accuracy.]{\includegraphics[width=\textwidth]{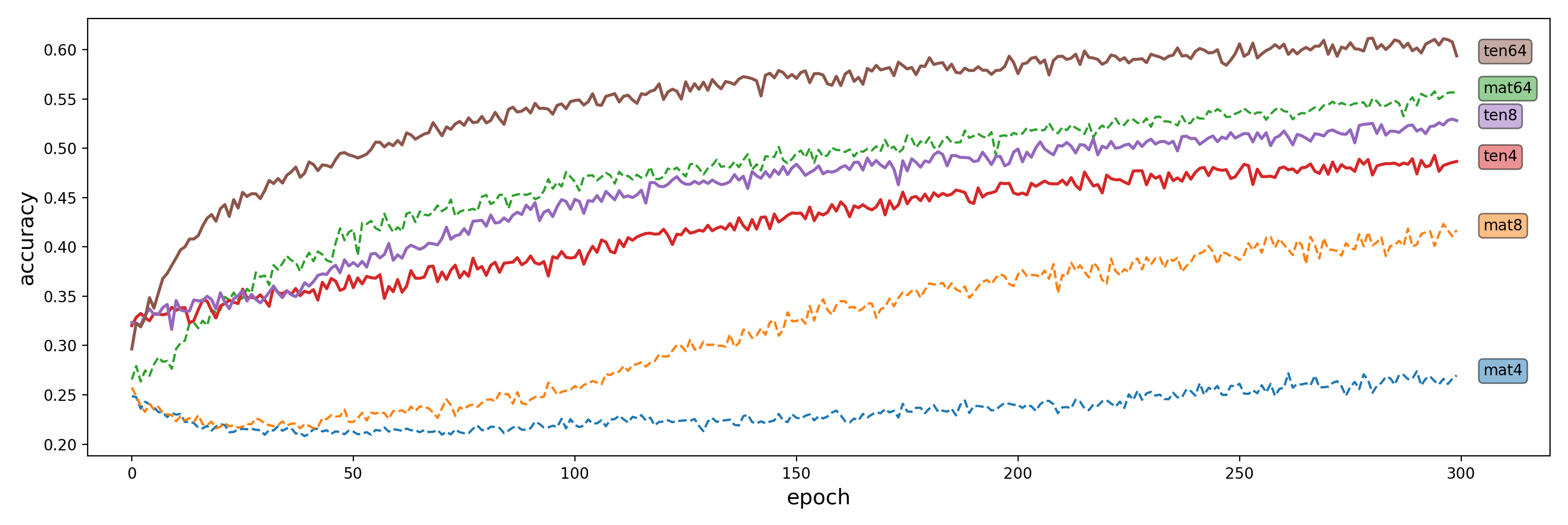}}
	
	\subfigure[CIFAR-10 loss.]{\includegraphics[width=\textwidth, height=5cm]{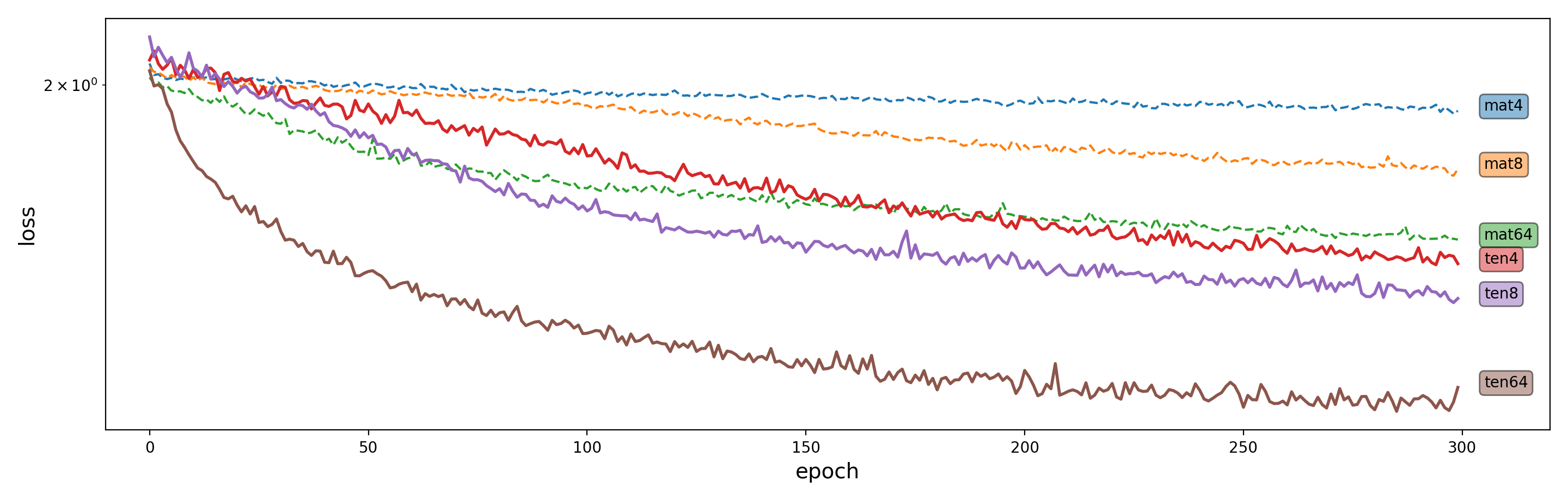}}
	\caption{CIFAR-10 results comparing matrix and tensor networks with $4$, $8$, and $64$ leapfrog layers.}
	\label{fig:cifar10 results}
\end{figure}

We compare our $t$-NN with leapfrog integration \cref{eqn:tensor leapfrog} to the matrix equivalent on both the MNIST dataset \cite{MNIST} and the CIFAR-10 dataset \cite{CIFAR10}.   We implement both tensor and matrix leapfrog frameworks using Pytorch \cite{Pytorch}, a deep-learning library with automatic differentiation and ample GPU acceleration.  We employ a stochastic gradient descent optimizer with a fixed learning rate of $0.1$ for the MNIST experiments and $0.01$ for the CIFAR-10 experiments and a momentum set to $0.9$.  We evaluate our performance with a cross-entropy loss function (\emph{tensor cross-entropy} in the case of $t$-NNs).  
We use a step size of $h = 0.1$ in the leapfrog discretization.  For the tensor networks, we use the $M$-product where $M$ is the orthogonal discrete cosine transform matrix.  In \cref{fig:mnist results} and \cref{fig:cifar10 results}, we plot the accuracy and loss of the network of the test data per epoch for various depths of the network.

The MNIST dataset is composed of $28\times 28$ grayscale images of handwritten digits, $60000$ training and $10000$ test images.  The images are centered about the center of mass, and we normalize the images to have a mean of $0.1307$ and a standard deviation of $0.0381$, as suggested by the Pytorch tutorials \cite{Pytorch}.  In stochastic gradient descent, we use a training and test batch size of $100$. For the matrix case, we vectorize each image into a $784 \times 1$ vector and apply square weight matrices of size $784 \times 784$ to preserve the number of output features at each layer.  For the tensor case, we orient each image as a lateral slice of size $28 \times 1 \times 28$ and apply cube weight tensors of size $28\times 28\times 28$.

We notice that both the matrix and tensor leapfrog networks converge to a high accuracy of $97$-$98\%$, even though the tensor network has an order of magnitude fewer weight parameters.  The tensor advantage is most apparent when comparing the tensor network with $4$ leapfrog layers to the matrix network with $8$ leapfrog layers.  The tensor network performs nearly as well as the matrix case, despite having significantly fewer learnable parameters and a shallower network.

The convergence behavior of the loss functions is even more telling than the convergence of the accuracy.  Using a tensor loss function, we obtain a more rapid and greater decrease the loss evaluation than in the matrix cases with a traditional cross-entropy loss function.  The rapid descent of the loss, particularly in the $8$-layer $t$-NN, demonstrates the efficiency of fitting our model using our tensor framework.  We are able to fit our model quickly while maintaining high accuracy, and we exhibit a greater improvement in our model as we update our parameters.  In the matrix networks, the loss quickly stagnates, and this enables the $t$-NNs to exceed the accuracy of the matrix networks.


The CIFAR-10 dataset is composed of $32 \times 32 \times 3$ RGB images belonging to $10$ distinct classes, $50000$ training and $10000$ test images and each set evenly split between different classes.  We normalize the (R,G,B) channels of each image to have means of $(0.4914, 0.4822, 0.4465)$ and standard deviations of $(0.2023, 0.1994, 0.2010)$, as suggested by \cite{NormalizationCIFAR10}.  In stochastic gradient descent, we use a training batch size of $128$ and a test batch size of $100$.  For the matrix case, we vectorize each image into a $3072 \times 1$ vector and apply square weight matrices of size $3072 \times 3072$.  For the tensor case, we concatenate the channel images along the first dimension thereby storing each image as a lateral slice of size $96 \times 1\times 32$ and apply weight tensors of size $96 \times 96 \times 32$.

The convergence behavior exhibited in \cref{fig:cifar10 results} is rather striking.  Despite having fewer learnable parameters, the $t$-NNs exhibit superior accuracy and loss results to their matrix counterparts.  Because of the more powerful $t$-NN parameterization, we converge to our top accuracy more rapidly and the loss converges more quickly to a lower value, indicating our $t$-NN model is extracting more significant features from the original data.  We are able to achieve 60\% accuracy without using convolutions, only using $t$-linear layers.

 Even more surprising, with a significantly shallower network, $t$-NNs with $8$ leapfrog layers produce comparable results to matrix networks with $64$ leapfrog layers.  This is evidence that tensor-tensor products not only enable rapid convergence  due to the reduced number of parameters, but also have the ability to encode more meaningful features; i.e., a more powerful parameterization.  
 One potential justification for our ability to learn with a shallower network is because we incorporate \emph{$t$-linear} rather than linear operators, therefore we need less non-linearity (i.e., fewer non-linear layers) to capture the same level of network complexity.

\section{Conclusions}\label{sec:conclusion}


We have shown that our t-product-based tensor neural network is a natural multidimensional extension of traditional neural networks.  In this high-dimensional framework, we have the potential to encode more information with fewer parameters.  Furthermore, inspired by recent theoretical advances connecting matrix-based neural networks to notions of stability in PDE discretizations, we have shown that our network supports a stable tensor forward propagation scheme which provides a more robust classification framework.  
We can also improve the robustness of our algorithm by regularizing our parameters to promote smooth dynamics of the data during forward propagation \cite{HaberRuthotto2017}. 
%


Going forward, we plan to test our stable $t$-NN on more challenging datasets and examine its robustness to adversarial attacks \cite{Papernot2016}.  Additionally, we are interested in improving our overall performance by using convolutions and creating a tensor convolutional neural network \cite{Krizhevsky2012,He2015}.  We also would like to explore the efficacy of using a variety of tensor-tensor products, and ultimately develop the product from the data itself.      

%
%
%
%

\section*{Acknowledgements}

We thank Eldad Haber and Lars Ruthotto for providing additional insight on their work which helped us generalize their ideas to our high-dimensional framework.

\appendix

\section{Tensor loss function back-propagation}\label{sec:tensor loss function backprop}

Below is a derivation of the back-propagation of the tubal softmax loss function.

\begin{derivation}[Tensor loss function back-propagation]\label{deriv:tensor softmax backprop}

\normalfont

We will derive the tensor loss back-propagation formula for a single training sample stored as a lateral slice; the formula naturally generalizes to multiple training samples.
 Let $\vec{\Acal}_N \in \Rbb^{\ell_N \times 1\times n}$ to be the network output of a single training sample, and suppose we apply the tensor softmax function described in \cref{eqn:softmax tubal} and \cref{eqn:softmax scalar} to obtain a vector of probabilities $\vec{y} = f(\Wcal_N * \vec{\Acal}_N)$.
We evaluate the performance of our network on $\vec{\Acal}_N$ in an objective function $\bs \Fcal = \bs \Lcal(\vec{y}, \vec{c})$\footnote{A typical loss function is cross-entropy loss \cite{Nielsen2017}.}.  To improve our performance via back-propagation, we first compute the error in our performance due to the output from our network, denoted as follows:
	\begin{align}\label{eqn:loss function chain rule1}
	\frac{\partial \bs \Fcal}{\partial \vec{\Acal}_N} 
		= \frac{\partial \bs \Lcal}{\partial \vec{y}} \cdot \frac{\partial \vec{y}}{\partial  \vec{\Acal}_N}
		= \frac{\partial \bs \Lcal}{\partial \vec{y}} \cdot \frac{\partial f(\Wcal_N * \vec{\Acal}_N)}{\partial  \vec{\Acal}_N}.
	\end{align}

For notational simplicity, let $\vec{\Xcal} = \Wcal_N * \vec{\Acal}_N$ and $\vec{\Ycal} = h(\vec{\Xcal})$.  Then, expressing $f(\vec{\Xcal}) = \texttt{sum}(\vec{\Ycal}, 3)$, the expanded multivariable chain rule is the following:
	\begin{align}\label{eqn:loss function chain rule2}
	\frac{\partial \bs \Lcal}{\partial \vec{y}} \cdot \frac{\partial \texttt{sum}(\vec{\Ycal},3)}{\partial  \vec{\Acal}_N}
	=\frac{\partial \bs \Lcal}{\partial \vec{y}} \cdot \frac{\partial \texttt{sum}(\vec{\Ycal},3)}{\partial \vec{\Ycal}}
		\cdot \frac{\partial \vec{\Ycal}}{\partial \vec{\Xcal}} \cdot \frac{\partial \vec{\Xcal}}{\partial \vec{\Acal}_N}.
	\end{align}

The key differentiation step is $\partial \vec{\Ycal} / \partial \vec{\Xcal}$, where we differentiate the tubal softmax function $h$.  Because we apply tubal functions tube-wise in \cref{eqn:softmax tubal}, we differentiate tube-wise as follows:
	\begin{align}\label{eqn:softmax tubal derivative}
	\frac{\partial \bs y_i}{\partial \bs x_j} 
		&= \frac{\partial}{\partial \bs x_j} \left[ (\sum_{j=1}^p \texttt{exp}(\bs x_j))^{-1} * \texttt{exp}(\bs x_i)\right]
	\end{align}
	

The derivation of the softmax back-propagation formula is rather cumbersome and not particularly enlightening.  Instead, we derive the derivative of the exponential tubal function, which will shed some light on how to complete the derivation for the softmax function.  Suppose we wish to differentiate the following, which we rewrite using the same intuition as \cref{eqn:tprod function apply eigenvalues}:
	\begin{align}\label{eqn:exponent derivative}
	\frac{\partial}{\partial \bs x}(\texttt{exp}(\bs x)) \equiv \frac{\partial}{\partial \vec{x}}(F^H \texttt{exp}(F\vec{x})),
	\end{align}
	where $F$ is the (normalized) DFT matrix, $\vec{x} = \texttt{vec}(\bs x)$, and $\texttt{exp}(\cdot)$ is applied element-wise in the transform domain.  We can now differentiate \cref{eqn:exponent derivative}; however, there is a small caveat that the matrix $F$ are complex, and hence we apply the conjugate transpose when we differentiate \cite{MatrixCookbook}.  The derivative in the transform domain is the following:
	\begin{align}\label{eqn:exponent derivative2}
	\frac{\partial}{\partial \vec{x}}(F^H \texttt{exp}(F\vec{x})) &= F^H(\texttt{exp}(F\vec{x})\odot (F\partial \vec{x})),
	\end{align}
	where $\delta \vec{x}$ comes from the chain rule.  Rewriting \cref{eqn:exponent derivative2} in terms of the $t$-product, we obtain the following formula:
	\begin{align}
	F^H(\texttt{exp}(F\vec{x})\odot (F\partial \vec{x})) \equiv \texttt{exp}(\bs x) * \partial \bs x.
	\end{align}
	This is exactly the formula we would expect, further demonstrating the matrix-mimetic properties of our tensor algebra. The derivation of the softmax back-propagation formula requires a similar process of differentiating in the transform domain.

\end{derivation}

\bibliographystyle{siamplain}
\bibliography{references}

\begin{thebibliography}{10}

\bibitem{Ascher2010}
{\sc U.~M. Ascher}, {\em Numerical methods for evolutionary differential
  equations}, SIAM,  (2010).

\bibitem{Atkinson1989}
{\sc K.~E. Atkinson}, {\em An Introduction to Numerical Analysis}, Wiley,
  second~ed., 1989.

\bibitem{Bengio1994}
{\sc Y.~Bengio, P.~Simard, and P.~Frasconi}, {\em Learning long-term
  dependencies with gradient descent is difficult}, IEEE Transactions on Neural
  Networks, 5 (1994), pp.~157--166.

\bibitem{Brooks2011}
{\sc S.~Brooks, A.~Gelman, G.~L. Jones, and X.-L. Meng}, {\em Handbook of
  Markov Chain Monte Carlo}, Chapman and Hall/CRC, 2011.

\bibitem{Chien2018}
{\sc J.-T. Chien and Y.-T. Bao}, {\em Tensor-factorized neural networks}, IEEE
  Transactions on Neural Networks, 29 (2018).

\bibitem{Denil2014}
{\sc M.~Denil, B.~Shakibi, L.~Dinh, M.~A. Ranzato, and N.~de~Frietas}, {\em
  Predicting parameters in deep learning}, arXiv:1306.0543v2,  (2014).

\bibitem{Goodfellow2016}
{\sc I.~Goodfellow, Y.~Bengio, and A.~Courville}, {\em Deep learning}, MIT
  Press,  (2016).

\bibitem{HaberRuthotto2017}
{\sc E.~Haber and L.~Ruthotto}, {\em Stable architectures for deep neural
  networks}, arXiv:1705.03341,  (2017).

\bibitem{HaberRuthottoHolthamJun2017}
{\sc E.~Haber, L.~Ruthotto, E.~Holtham, and S.-H. Jun}, {\em Learning across
  scales - multiscale methods for convolution neural networks},
  arXiv:1703.02009,  (2017).

\bibitem{Hao2014}
{\sc N.~Hao, L.~Horesh, and M.~Kilmer}, {\em Nonnegative tensor decomposition},
  Compressed Sensing \& Sparse Filtering, Springer, 2014, pp.~123--148.

\bibitem{HaoKilmer2013}
{\sc N.~Hao, M.~E. Kilmer, K.~Braman, and R.~C. Hoover}, {\em Facial
  recognition using tensor-tensor decompositions}, SIAM Journal of Imaging
  Sciences, 6 (2013), pp.~437--463.

\bibitem{He2015}
{\sc K.~He, X.~Zhang, S.~Ren, and J.~Sun}, {\em Deep residual learning for
  image recognition}, arXiv:1512.03385,  (2015).

\bibitem{Hinton2012}
{\sc G.~Hinton, L.~Deng, D.~Yu, G.~Dahl, A.~rahman Mohamed, N.~Jaitly,
  A.~Senior, V.~Vanhoucke, P.~Nguyen, T.~N. Sainath, and B.~Kingsbury}, {\em
  Deep neural networks for acoustic modeling in speech recognition: The shared
  views of four research groups}, IEEE Signal Processing Magazine, 29 (2012).

\bibitem{KernfeldKilmer2015}
{\sc E.~Kernfeld, M.~Kilmer, and S.~Aeron}, {\em Tensor-tensor products with
  invertible linear transforms}, Linear Algebra and its Applications, 485
  (2015), pp.~545--570.

\bibitem{KilmerBraman2013}
{\sc M.~E. Kilmer, K.~Braman, N.~Hao, and R.~C. Hoover}, {\em Third-order
  tensors as operators on matrices: a theoretical and computational framework
  with applications in imaging}, SIAM Journal on Matrix Analysis and
  Applications, 34 (2013), pp.~148--172.

\bibitem{KilmerMartin2011}
{\sc M.~E. Kilmer and C.~D. Martin}, {\em Factorization strategies for
  third-order tensors}, Linear Algebra and its Applications, 435 (2011),
  pp.~641--658.

\bibitem{CIFAR10}
{\sc A.~Krizhevsky}, {\em Learning multiple layers of features from tiny
  images}, 2009.

\bibitem{Krizhevsky2012}
{\sc A.~Krizhevsky, I.~Sutskever, and G.~E. Hinton}, {\em Imagenet
  classification with deep convolutional imagenet classification with deep
  convolutional neural networks}, Proceedings of the 25th International
  Conference on Neural Information Processing Systems,  (2012), pp.~1091--1105.

\bibitem{LeCun1998}
{\sc Y.~LeCun, L.~Bottou, Y.~Bengio, and P.~Haffner}, {\em Gradient-based
  learning applied to document recognition}, Proceedings of the IEEE,  (1998).

\bibitem{MNIST}
{\sc Y.~LeCun, C.~Cortes, and C.~J. Burges}, {\em The mnist database of
  handwritten digits}, 1998.

\bibitem{NormalizationCIFAR10}
{\sc K.~Liu}, {\em 95.16\% on cifar10 with pytorch}.
\newblock https://github.com/kuangliu/pytorch-cifar, 2018.

\bibitem{Lund2018}
{\sc K.~Lund}, {\em The tensor t-function: A defintion for functions of
  third-order tensors}, arXiv:1806.07261v1,  (2018).

\bibitem{Martin2013}
{\sc C.~D. Martin, R.~Shafer, and B.~Larue}, {\em An order-p tensor
  factorization with applications in imaging}, SIAM Journal of Scientific
  Computing, 35 (2013), pp.~A474--A490.

\bibitem{Mhaskar2016}
{\sc H.~Mhaskar and T.~Poggio}, {\em Deep vs. shallow networks: An
  approximation theory perspective}, Analysis and Applications, 14 (2016),
  pp.~829--848.

\bibitem{Newman2017}
{\sc E.~Newman, M.~Kilmer, and L.~Horesh}, {\em Image classification using
  local tensor singular value decompositions}, in CAMSAP, 2017.

\bibitem{Nielsen2017}
{\sc M.~Nielsen}, {\em Neural networks and deep learning}.
\newblock http://neuralnetworksanddeeplearning.com/, 2017.

\bibitem{Novikov2015}
{\sc A.~Novikov, D.~Podoprikhin, A.~Osokin, and D.~Vetrov}, {\em Tensorizing
  neural networks}, arXiv:1509.06569v2,  (2015).

\bibitem{Papernot2016}
{\sc N.~Papernot, P.~McDaniel, M.~Fredrikson, Z.~B. Celik, and A.~Swami}, {\em
  The limitations of deep learning in adversarial settings}, IEEE European
  Symposium on Security and Privacy,  (2016).

\bibitem{Pytorch}
{\sc A.~Paszke, S.~Gross, S.~Chintala, G.~Chanan, E.~Yang, Z.~DeVito, Z.~Lin,
  A.~Desmaison, L.~Antiga, and A.~Lerer}, {\em Automatic differentiation in
  pytorch}, in NIPS-W, 2017.

\bibitem{MatrixCookbook}
{\sc K.~B. Petersen and M.~S. Pedersen}, {\em The Matrix Cookbook}, Technical
  University of Denmark, November 2012.

\bibitem{Phan2010}
{\sc A.~H. Phan and A.~Cichocki}, {\em Tensor decompositions for feature
  extraction and classification of high dimensional datasets}, Nonlinear theory
  and its applications, IEICE, 1 (2010), pp.~37--68.

\bibitem{Rumelhart1986}
{\sc D.~E. Rumelhart, G.~E. Hinton, and R.~J. Williams}, {\em Learning
  representations by back-propagating errors}, Nature, 323 (1986),
  pp.~533--536.

\bibitem{ShalevShwartz2017}
{\sc S.~Shalev-Shwartz, O.~Shamir, and S.~Shammah}, {\em Failures of
  gradient-based deep learning}, arXiv:1703.07950,  (2017).

\bibitem{SoltaniKilmerHansen2016}
{\sc S.~Soltani, M.~E. Kilmer, and P.~C. Hansen}, {\em A tensor-based
  dictionary learning approach to tomographic image reconstruction}, BIT
  Numerical Mathematics,  (2016).

\bibitem{Stoudenmire2016}
{\sc E.~Stoudenmire and D.~Schwab}, {\em Supervised learning with tensor
  networks}, NIPS,  (2016).

\bibitem{Zhang2014}
{\sc Z.~Zhang, G.~Ely, S.~Aeron, N.~Hao, and M.~Kilmer}, {\em Novel methods for
  multilinear data completion and denoising based on tensor-svd}, CVPR,
  (2014).

\end{thebibliography}

\end{document}